\documentclass{ecai} 

\usepackage{latexsym}
\usepackage{amssymb}
\usepackage{amsmath}
\usepackage{amsthm}
\usepackage{booktabs}
\usepackage{enumitem}
\usepackage{graphicx}
\usepackage{color}
\usepackage{algorithm}
\usepackage{algorithmic}
\usepackage{stfloats}
\usepackage{siunitx}
\usepackage{epsfig}
\usepackage{float}
\usepackage{hyperref}
\usepackage[justification=justified]{caption}

\newcommand{\BibTeX}{B\kern-.05em{\sc i\kern-.025em b}\kern-.08em\TeX}

\begin{document}


\begin{frontmatter}


\paperid{206} 


\title{An Animation-based Augmentation Approach for Action Recognition from Discontinuous Video}


\author{\fnms{Xingyu}~\snm{Song}}
\author{\fnms{Zhan}~\snm{Li}}
\author{\fnms{Shi}~\snm{Chen}} 
\author{\fnms{Xin-Qiang}~\snm{Cai}} 
\author{\fnms{Kazuyuki}~\snm{Demachi}} 

\address{The University of Tokyo}
\address{songxingyu0429@gmail.com, \{lizhan, shichen, yypr9411\}@g.ecc.u-tokyo.ac.jp, cai@ms.k.u-tokyo.ac.jp}


\begin{abstract}
Action recognition, an essential component of computer vision, plays a pivotal role in multiple applications. 
Despite significant improvements brought by Convolutional Neural Networks (CNNs), these models suffer performance declines when trained with discontinuous video frames, which is a frequent scenario in real-world settings. 
This decline primarily results from the loss of temporal continuity, which is crucial for understanding the semantics of human actions. 
To overcome this issue, we introduce the 4A (Action Animation-based Augmentation Approach) pipeline, which employs a series of sophisticated techniques: starting with 2D human pose estimation from RGB videos, followed by Quaternion-based Graph Convolution Network for joint orientation and trajectory prediction, and Dynamic Skeletal Interpolation for creating smoother, diversified actions using game engine technology. 
This innovative approach generates realistic animations in varied game environments, viewed from multiple viewpoints. 
In this way, our method effectively bridges the domain gap between virtual and real-world data. 
In experimental evaluations, the 4A pipeline achieves comparable or even superior performance to traditional training approaches using real-world data, while requiring only 10\% of the original data volume. 
Additionally, our approach demonstrates enhanced performance on In-the-wild videos, marking a significant advancement in the field of action recognition.
Code and model are available at \href{https://github.com/xingyu-song/4A}{here}.
\end{abstract}

\end{frontmatter}


\section{Introduction}
Action recognition is a critical component of computer vision that involves identifying and classifying various actions from sequences of images or video frames. 
This task is essential across numerous applications, including malicious behavior identification, accident detection, and human-computer interaction~\cite{LI2024121367, intro_safe}. 
The significant advancements in Convolutional Neural Networks (CNNs) notably enhances performance in action recognition tasks across a range of benchmark datasets~\cite{dataset05_Kinetics, model01_TSN, ntu, dataset01_HVU}. 

However, when encountering scenarios with discontinuous frame sequences during training, which is a common occurrence in real-world settings, CNN-based models implemented for action recognition suffer in a significant performance decline. 
For instance, with continuous frame training videos achieving around 40\% mean accuracy, while training with missing frames drops to below 20\% (refer to Section~\ref{sec:results} for details).

Our motivation for addressing discontinuous videos during training is grounded in real-world challenges where data collection is often incomplete due to factors like occlusions, bandwidth limitations, and privacy concerns. In environments such as surveillance, sports analytics, and remote monitoring in hazardous areas, video recordings are frequently captured in non-continuous segments rather than a single, uninterrupted stream. 

Unlike other CNN-implemented tasks related to human motion, such as {pose estimation}~\cite{pe_glagcn, 3D_PE_semi, pr_semgcn, 3D_PE_semi} or {3D human reconstruction}~\cite{render4cnn, shape_SMPLx, 3d_rec_hmmr}, \textbf{do not} exhibit such severe performance declines with discontinuous frames~\cite{surreal}.  
For instance,~\cite{jointformer} demonstrates less than 1\% of drop in performance when training on the dataset with extracted frames compared to training on the original frame sequence from H3WB dataset~\cite{h3wb}.
Unlike these tasks, action recognition fundamentally involves a deeper \textbf{semantic} analysis. 
It requires the interpretation and understanding of human motion patterns, essentially deciphering the meanings or semantics behind those actions~\cite{semantic1}. 
Therefore, the absence of temporal information due to missing frames directly diminishes the understanding of an action, making the action recognition task susceptible to the continuity of the video. 
On the other hand, the loss of semantics from the original data complicates the process of augmenting the dataset as well.

\begin{figure*}[ht]
\includegraphics[width=\textwidth]{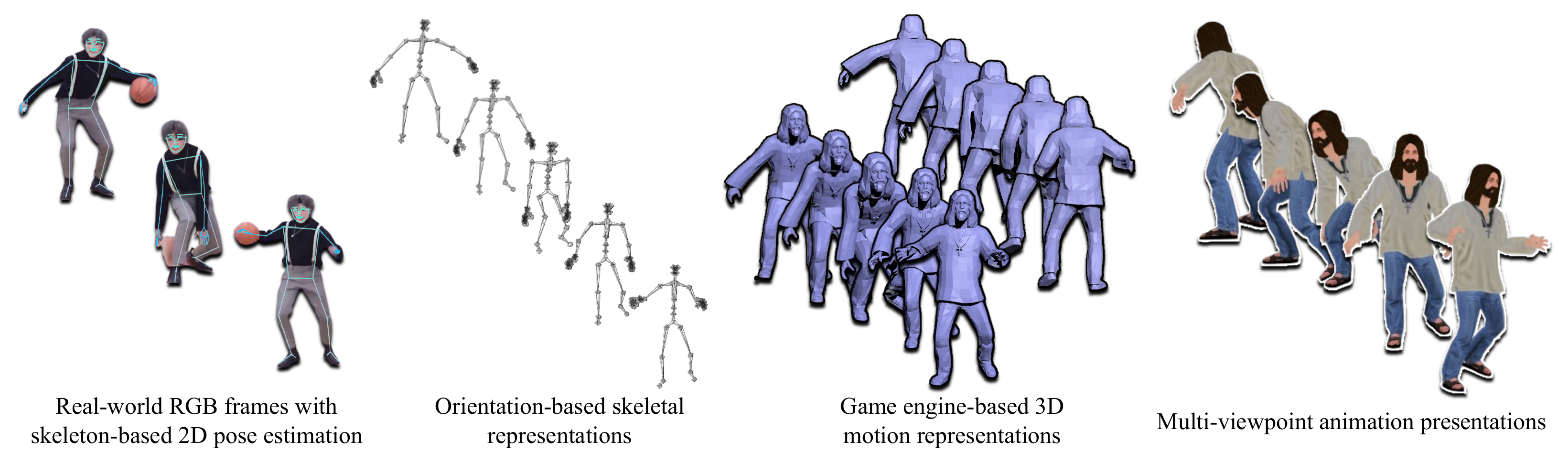} 
\vspace{2pt}
\caption{Overview of 4A pipeline. Within 4A pipeline, we begin with a 2D human pose estimation method to extract the 2D coordinates of human skeleton coordinates from real-world RGB videos.
This is followed by employing a Quaternion-based Graph Convolution Network (Q-GCN) to predict the orientation of each bone joint and the trajectory of body in 3D space.
Subsequently, the Dynamic Skeletal Interpolation algorithm (DSI) ensures a smoother and more diversified action animation.
After that, we use the game engine technology to generate the motion from skeleton representation sequence to form the animation.
Finally, we present the animation in game environment with diverse environments and appearances, and captured in multiple viewpoints. }
\label{fig:overall}
\end{figure*}

Inspired by the previous studies on data augmentation in other computer vision tasks~\cite{surreal, render4cnn, synthetic_humans}, we propose to use synthetic human to mitigate the issue of missing frames in action recognition tasks.
In this study, we introduce the 4A (Action Animation-based Augmentation Approach) pipeline, an innovative, efficient, and scalable pipeline for data augmentation within the action recognition field. 
This approach achieves the generation of smooth and realistic (natural-looking) synthetic human motions (termed animations), depicted across a variety of settings, appearances, and conditions from multiple viewpoints, leveraging discontinues monocular RGB videos from real world. 
The detailed pipeline of 4A is illustrated in Figure~\ref{fig:overall}. 
Furthermore, we conduct experiments to evaluate the effectiveness of 4A in bridging the domain gap between virtual representations and real-world tasks.

The main contributions of our work include:
(1) we discover the problem of severe decrease on performance of action recognition task training by discontinuous video, and the limitation of existing augmentation methods on solving this problem. 
(2) we propose a novel augmentation pipeline, 4A, to address the problem of discontinuous video for training, while achieving a smoother and much more natural-looking action representation than the latest data augmentation methodology. 
(3) We achieve the same performance with only 10\% of the original data for training as with all of the original data from the real-world dataset, and a better performance on In-the-wild videos, by employing our data augmentation techniques.

\section{Related Work}
\subsection{Synthetic Human Construction} 

Recent advancements have seen synthetic images of humans deployed to train visual models for tasks like 2D or 3D body pose and shape estimation~\cite{relate_sysshape_2}, part segmentation~\cite{relate_poseseg_1, surreal}, and person re-identification~\cite{relate_reid}. 
However, synthetic datasets built for these tasks often lack action labels, limiting their applicability for action recognition tasks.

Prior researches leveraging synthetic human data for action recognition are few~\cite{relate_ar_1, relate_ar_3}, with some researches focusing on synthetic 2D human pose sequences~\cite{relate_ar_4} and point trajectories~\cite{relate_ar_5} for view-invariant action recognition. 
RGB-based synthetic training data for action recognition is a field under exploration, with few attempts addressing the manual definition of action classes for multitask learning~\cite{relate_ar_1}. 
However, scalability and relevance to target classes remain challenges in these approaches.

A study more aligned with our work~\cite{relate_ar_3} uses synthetic training images derived from RGB-D inputs to enhance performance on unseen viewpoints, framing a pose classification problem that serves as a basis for action recognition. 
Yet, the discriminative power of these features for specific action categories is questionable. 
In contrast, another approach~\cite{synthetic_humans} extracts motion sequences directly from real data, offering flexibility for incorporating new categories and assigning explicit action labels to synthetic videos. 
This method, however, suffers from mismatches between characters and their environments, alongside issues with action smoothness and photorealism.
This is primarily due to the limitations of 3D human shape estimation or reconstruction technologies~\cite{shape_SMPLx, 3d_rec_hmmr}, especially in handling videos with discontinuous frames.


\subsection{Game Engine-based Action Datasets.} 
Video game-based datasets have been increasingly utilized for training deep learning models. 
JTA~\cite{jta}, for example, is a vast dataset created using a video game for pedestrian pose estimation and tracking in urban environments. 
GTA-IM~\cite{gamedataset02_GTAIM}, a pose estimation dataset, highlights human-scene interactions and employs a developed game engine interface for automatic control of characters, cameras, and actions. 
However, both them focus on static human poses, rather than capturing temporal movements with semantics. 
NCTU-GTA360~\cite{gamedataset04_GTA360}, an action recognition dataset featuring spherical projection captured from video games, involves the use of 360-degree cameras to record the entire surroundings of a character.
Nevertheless, NCTU-GTA360 faces an imbalance in action distribution, with basic actions like ``OnFoot'' or ``Stopped'' overwhelmingly dominating more complex actions such as ``Ragdoll'' or ``SwimmingUnderWater''. 
Other datasets like SIM4ACTION~\cite{gamedataset03_SIM} and G3D~\cite{gamedataset01_G3D} also share a common limitation in offering a constrained range of action classes. 
Most notably, none of these datasets, including GTA-IM and NCTU-GTA360, have succeeded in effectively importing a vast amount of self-customized actions from the real world, which is crucial for creating more comprehensive and diverse training models.

\subsection{3D Skeletal Motion Representation} 
\label{3D Skeletal Motion Representation}

3D Human Pose Estimation (HPE) in videos, aiming to predict human body joint locations in 3D space, employs methods like single-stage and 2D to 3D lifting. 
While single-stage methods estimate 3D pose directly from images~\cite{ma2021context, RN021}, 2D to 3D lifting~\cite{3D_PE_semi,RN013}, using ground truth 2D poses, generally performs better. 
However, both approaches face challenges in providing smooth motion representations from discontinuous frames, indicating a gap in effectively handling interrupted sequences for accurate 3D skeletal motion representation.

\subsection{3D Human Shape Estimation}
\label{3D Human Shape Estimation}
The Skinned Multi-Person Linear model (SMPL)~\cite{shape_SMPLx} represents a pivotal approach in human motion capture, marking the first introduction of orientation-based human body representation. 
Subsequent developments inspired by SMPL, including models like MANO~\cite{shape_MANO:SIGGRAPHASIA:2017}, SMPL-X~\cite{shape_SMPLx}, and STAR~\cite{shape_STAR:2020}, have expanded the framework's utility to encompass detailed body shape modeling, facial expressions, hand movements, and the representation of clothed human bodies. 
Despite these advancements, similar to challenges in 3D pose estimation, the performance of human shape estimation models is constrained by their ability to express the semantics of motion, particularly when trained on limited datasets.

\section{The 4A Pipeline}
The goal of 4A pipeline is to improve the performance of action recognition using synthetic data especially when the training videos are discontinuous or insufficient. 
There are four stages in 4A:
(1) 2D skeleton extraction;
(2) 3D orientation lifting;
(3) Sequence smoothing;
(4) Animation generation and capturing;
\subsection{2D Skeleton Extraction}
\label{2D Skeleton Extraction}
In this stage, we utilize a 2D human skeleton estimation technique, HRNet~\cite{hrnet}, trained on COCO-WholeBody dataset~\cite{coco_whole-body}, to extract 2D human wholebody skeleton keypoints from the RGB frames of monocular videos. 
For more semantic representation, we construct a hierarchical structure of human skeleton following the biovision hierarchy file format inspired by~\cite{CDGCN}. 
However, unlike~\cite{CDGCN}, which only contains 17 joint nodes and 16 bone joints, our configuration contains 54 node joints and 53 bone joints to form a more detailed whole-body motion including hands, feet and neck. 
Furthermore, the upper and lower body respectively form an hierarchical inheritance structure (start from pelvis), extending from the parent joint to the child joint of both each node joint and bone joint, where the bone joint can be regraded as a vector while the node joint is the start and end point of it.

\subsection{3D Orientation Lifting}
In this stage, we attempt to lift 2D skeletal representation into 3D space to present motion in multiple viewpoints. 
However, due to the unsatisfying performance of 3D pose and shape estimation (mentioned in Section~\ref{3D Skeletal Motion Representation} and~\ref{3D Human Shape Estimation}), which are unable to provide viable motion representations in 3D space. 
It is important to note that, unlike 3D pose estimation, the reconstruction of synthetic human motion does not require precise joint coordinates; approximate dynamics of human motion are sufficient.
Drawing inspiration from the SMPL model~\cite{shape_SMPLx}, we extend our hierarchical human skeleton structure by using the coordinates of the root node (pelvis) and the orientations of other bone joints to form the skeleton representation in 3D space in each frame. 
Consequently, predicting the 3D space orientation of each bone joint becomes our primary challenge. 

Prior research utilizing Graph Convolution Network (GCN) for human pose estimation~\cite{pe_glagcn, 3D_PE_semi} and action recognition~\cite{model12_stgcn, ar_ddgcn, DGCN} demonstrates the powerful capability of GCNs in extracting human dynamic features. 
Influenced by the techniques in~\cite{CDGCN} and~\cite{3D_PE_semi}, we develop a Quaternion Graph Convolution Network (Q-GCN) to predict the Quaternions (used to representing orientation) and the root 3D coordinates for each bone joint from 2D coordinates of each skeleton keypoint.

\subsubsection{Q-GCN}

Similar as~\cite{DGCN, CDGCN, model12_stgcn, ar_ddgcn, pr_semgcn}, we first construct a graph of human body skeleton following our structure. 
The vertices of these graphs comprise the sequence of human poses within the 2D coordinate space, denoted as 
$P_{2D}=\{ \mathbf{X}_{t,j} \in \mathbb{R}^2 | t=1,2,\ldots,T; j=1,2,\ldots,J\}$, 
where $\mathbf{X}_{t,j}$ represents the 2D coordinates of joint node $j$ at frame $t$. 
Here, $T$ and $J$ respectively signify the number of frames in the sequence and the number of joints in the human skeleton. 
Differ from prior implementations of GCNs, we also compile a sequence of rotations for each bone joint within the 2D coordinate space to constitute the edges of graph, expressed as 
$R_{2D}=\{ \mathbf{Z}_{t,b} \in \mathbb{R}^2 | t=1,2,3,\ldots,T; b=1,2,3,\ldots,B\}$, 
where $B$ represents the number of bone joints and $\mathbf{Z}_{t,b}$ includes a 2-tuple consisting of the cosine and sine values of the rotation angle for bone joint $b$ from the initial position in Local Coordinate System (LCS).
Note that the initial position in LCS of a bone joint is defined by its parent node and bone joint in the hierarchical structure, overlapping with the extension of parent bone joint from the parent node as origin point, where the rotation angle $\theta \in [-\pi, \pi ]$.

Formally, this temporal sequence of graphs is articulated as 
$\mathcal{G} = (\mathcal{V},\mathcal{E})$, 
where $\mathcal{V} = \{ v_{t,j} | t=1,2,\ldots,T; j=1,2,\ldots,J\}$, and $\mathcal{E} = \{ e_{t,b} | t=1,2,\ldots,T; b=1,2,3,\ldots,B\}$, are the sets of vertices and edges respectively. 
Note that the features of vertex $v_{t,j}$ and edge $e_{t,b}$ are initialized with their corresponding 2D coordinates $\mathbf{X}_{t,j}$ and rotation $\mathbf{Z}_{t,b}$.

Subsequently, we employ our Q-GCN to predict the sequence of Quaternions $Q_{4D} = \{\mathbf{Q}_{b} \in \mathbb{R}^4 |  b=1,2,3,\ldots,B\}$ and the root coordinate in the 3D space $\mathbf{P}_{root}\in \mathbb{R}^3$, serving as the representation of orientation.
Similar as rotations in 2D system, orientations in 3D spaces are also defined within LCS.
This design is employed to enhance the understanding of the internal dynamics and influence exerted from parent nodes to child nodes according to the previous research~\cite{CDGCN}. 

Similar as~\cite{model12_stgcn}, we first implement a basic spatial-temporal graph convolution block to extract the feature within the graph. 
We define a neighbor set $\mathcal{B}_{j}^{v}$ as a spatial graph convolutional filter for vertex $v_{t,j}$ while set $\mathcal{B}_{b}^e$ for edge $e_{t,b}$.
Inspired by~\cite{pe_glagcn}, both for vertex and edge filters, we define four distinct neighbor subsets: (1) self, (2) parent, and (3) child. 
Therefore, the kernel size $K$ is set to 3, corresponding to the 3 subsets. 
To implement the subsets, mappings $h_{t,j}^{v} \rightarrow \{ 0,\dots, K-1\}$ and $h_{t,b}^{e}\rightarrow \{ 0,\dots, K-1\}$ are used to index each subset with a numeric label. 
Therefore, this convolutional operations of vertex and edge can be written as 
\begin{equation}
    f_{out}^{v}(v_{t,j}) = \sum_{v_{t,n_{j}} \in \mathcal{B}_{j}^{v}} \frac{1}{Z_{t,n_{j}}} f_{in}^{v}(v_{t,n_{j}})W_{v}(h_{t,j}^{v}(v_{t,n_{j}}))
    \label{equ:1}
\end{equation}

\begin{equation}
    f_{out}^{e}(e_{t,b}) = \sum_{e_{t,n_{b}} \in \mathcal{B}_{b}^{e}} \frac{1}{Z_{t,n_{b}}} f_{in}^{e}(e_{t,n_{b}})W_{e}(h_{t,b}^{e}(v_{t,n_{b}}))
    \label{equ:2}
\end{equation}
where $f_{in}^{v}(v_{t,n_{j}}): v_{t,n_{j}} \rightarrow \mathbb{R}^2$ and $f_{in}^{e}(e_{t,n_{b}}): e_{t,n_{b}} \rightarrow \mathbb{R}^2$ denote the mappings that get the attribute feature of neighbor node joint $v_{t,n_{j}}$ and neighbor bone joint $e_{t,n_{b}}$ respectively. 
$Z_{t,n_{j}}$ and $Z_{t,n_{b}}$ is the normalization term that equal to the subset's cardinality. 
$W(h_{t,j}(v_{t,n_{j}}))$ and $W(h_{t,b}(v_{t,n_{b}}))$ are the weight functions of mapping $\mathcal{B}_{j}^{v}$ and $\mathcal{B}_{b}^e$ respectively, which are implemented by indexing a $(2,K)$ tensor. 
Within a pose frame, the determined graph convolution of a sampling strategy can be implemented by adjacent matrices of $J \times J$  for vertex and $B \times B$ for edge. 
Specifically, with $K$ spatial sampling strategies $\sum_{k=0}^{K-1} A_{k}^{v}$ for vertex and $\sum_{k=0}^{K-1} A_{k}^{e}$ for edge, Equation~\ref{equ:1} and~\ref{equ:2} can be transformed into the expressions using matrices into:
\begin{equation}
    {H}_{t}^{v} = \sum_{k=0}^{K-1} \Bar{A}_{k}^{v} {F}_{k}^{v} {W}_{k}^{v}
\end{equation}

\begin{equation}
    {H}_{t}^{e} = \sum_{k=0}^{K-1} \Bar{A}_{k}^{e} {F}_{k}^{e} {W}_{k}^{e}
\end{equation}
Where $\Bar{A}_{k} = \Lambda_{k}^{\frac{1}{2}} A_{k} \Lambda_{k}^{\frac{1}{2}}$ is the normalized adjacency matrix of $A_{k}$ both for vertex and edge, with its elements indicating whether a vertex $v_{t,n_{v}}$ or a edge $e_{t,n_{e}}$ is included in the neighbor subset.
Similar as~\cite{gcn}, $\Lambda_{k}^{ii} = \sum_{n}(\Bar{A}_{k}^{in}) + \alpha$ is a diagonal matrix with $\alpha$ set to $0.001$ to prevent empty rows. 
$W_{k}$ denotes the weighting function of Equation~\ref{equ:1} and~\ref{equ:2}, which is a weight tensor of the $1\times1$ convolutional operation. 
$F_{k}$ is the attribute features of all the neighbor joints sampled into the subset $k$.
Therefore, a convolution layer in Q-GCN is realized with a $1 \times T$ classical 2D convolution layer, where $T$ is the temporal kernel size that we set to 10. 
And the output of layer $H_{t}$ is both followed by a batch normalization layer and a ReLU layer and a dropout layer after them to form a convolutional block. 
In addition, a residual connection~\cite{resnet} is added as well. 

\begin{figure}[ht]
    \centering
    \includegraphics[width=\linewidth]{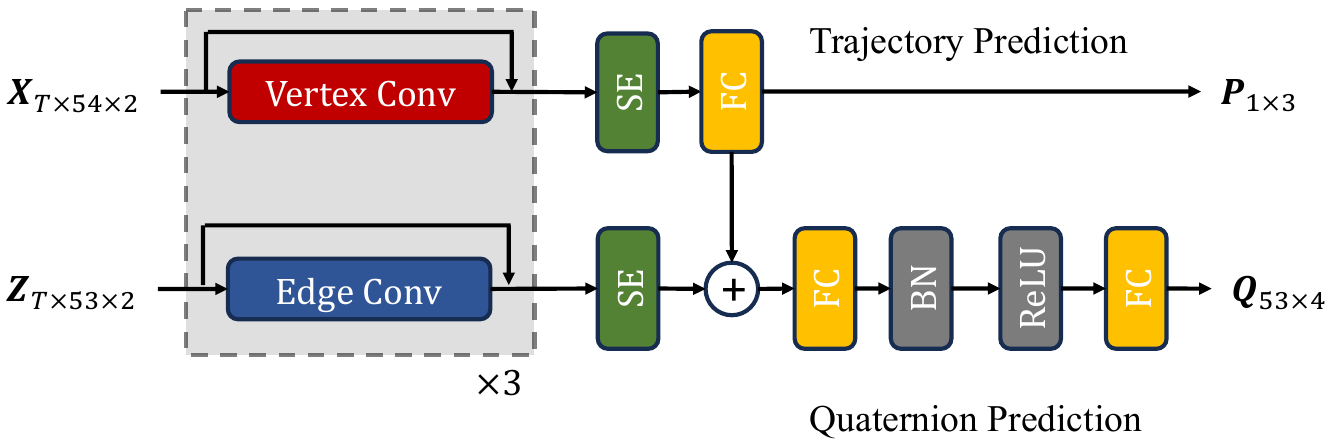}
    \vspace{2pt}
    \caption{Whole architecture of Q-GCN. It starts with three vertex and edge convolutional blocks, with residual connection operation in each block.
    After extract the neighbor features, both layers are followed by a Squeeze and Excitation (SE) Block.
    Then, the concatenation of vertex graph and edge graph is followed by two fully connection layer with batch normalization and ReLU function in between. }
    \label{fig:Q-GCN}
\end{figure}

The whole architecture of Q-GCN is shown in Figure~\ref{fig:Q-GCN}.
Inspired by~\cite{3D_PE_semi}, we also construct a trajectory prediction block to predict the global position of root node in a sequence. 
Finally, the Quaternion of each bone joint can also be predicted. 

As for the loss function for the trajectory prediction, we adopt the weighted mean per-joint position error~\cite{3D_PE_semi} (WMPJPE).
As for Quaternion prediction, we develop our Average Angular Distance (AAD) loss function to minimize the angular distance between the ground truth and predicated value: 
\begin{equation}
    \mathcal{L}_{angular} = \frac{1}{T} \frac{1}{B} \sum_{t=1}^{T} \sum_{b=1}^{B} 2\arccos{(Re(\boldsymbol{\Bar{Q}}_{t,b} \times conj(\boldsymbol{Q}_{t,b}))}
\end{equation}

Where $\mathbf{\Bar{Q}}_{t,b}$ and $\mathbf{Q}_{t,b}$ stand for the ground truth and the predicted value of Quaternion of bone joint $b$ at frame $t$.
And the functions $Re(\cdot)$ and $conj(\cdot)$ return the real part and the conjugate of a Quaternion respectively. 
In addition, we have also proposed a whole-body 3D orientation lifting dataset for training Q-GCN, derived from {H3WB}~\cite{h3wb}. 

\subsection{Sequence Smoothing}
After assembling the Quaternion of each bone joint for an individual frame, we proceed to compile them to create a continuous sequence across a stream of frames to depict a complete representation of action. 
However, directly combining Quaternion sequence will lead to frequent mismatches and jitters among the representation. 
Unlike typical time series data, directly applying a piecewise polynomial interpolation algorithm to orientation sequences can lead to significant issues, such as missing specific poses and decreased movement amplitude (detailed in Section~\ref{sec:results}). 

To tackle these challenges, we have developed the Dynamic Skeletal Interpolation (DSI) algorithm. 
This algorithm dynamically segments the Quaternion sequence into unit motions based on the variation in the range of motion. 
It then automatically interpolates the Quaternion sequence across different frame counts, ensuring a smoother and more natural-looking animation.
Finally, it randomly generates a series of variants for each sequence, ensuring a diverse representation.

The algorithm is detailed in Algorithm~\ref{alg}. 
Here, $\mathbf{Q}$ stands for the sequence of Quaternion, While $\mathbf{Q'}$ is the sequence after interpolation.
$d^{f}$ denoted the weighted average of the Angular Distance of each bone joint between adjacent frames, with $w_{b}$ representing the weight of each bone joint. 
The functions $Re(\cdot)$ and $conj(\cdot)$ return the real part and the conjugate of a Quaternion respectively. 
$p(x)_{[a,b]}$ refers to the Lagrange interpolating polynomial in the interval $[a,b]$, while $L(x)$ represents the Lagrange basis polynomial. 
$Linespace([a,b],n)$ is the function used to create evenly space numbers over interval $[a,b]$. 
The parameter $\delta$ is the interpolation rate, indicting the ratio of original frames to interpolated frames. 
$\eta$ is the interpolation coefficient.  
To enhance the diversity of an action, we develop Random Variation function $\mathcal{V}(\cdot)$ to generate a series of variants for each sequence.
To smooth the piecewise interpolated data, we employ Supersmoother~\cite{supersmoother} $\mathcal{S}(\cdot)$, a non-parametric smoothing method. 
$V, B, F, F'$ denote the number of variants, bone joints, frames before and after interpolation.

\begin{algorithm}[hb]
\caption{Dynamic Skeletal Interpolation}
\label{DSI}
\textbf{Input}: $\boldsymbol{Q} \in \mathbb{R}^{F \times B \times 4}$
\begin{algorithmic}[1] 
\STATE Define $\boldsymbol{Q}' \in \mathbb{R}^{F' \times B \times 4}$
\STATE Define $i=1$
\FOR{$f=2$ to $F$}
    \STATE $d^{f} = \frac{1}{J}\displaystyle\sum_{b=1}^B w_{b} \cdot 2\arccos{(Re(\boldsymbol{q}_{b}^{f} \times conj(\boldsymbol{q}_{b}^{f-1}))}$
    \IF{$d^{f}>threshold$}
        \STATE $p_{[i,f-1]}(x) = \displaystyle\sum_{b=1}^B w_{b} \sum_{t=i}^{f-1} \boldsymbol{q}_{b}^{t} \cdot L_{t}(x)$
        \STATE $p_{[f-1,f]}(x) = \displaystyle\sum_{b=1}^B w_{b}
        (\boldsymbol{q}_{b}^{f-1} \cdot L_{f-1}(x) + \boldsymbol{q}_{b}^{f} \cdot L_{f}(x))$
        \STATE $\boldsymbol{x}_{nor} = Linespace([i,f-1], \frac{1}{\delta})$
        \STATE $\boldsymbol{x}_{edge} = Linespace([f-1,f], Int(\frac{\eta \cdot d^{f}}{\delta}))$
        \STATE $\boldsymbol{Q}'_{[i,f-1]}=p_{[i,f-1]}(\boldsymbol{x}_{nor})$
        \STATE $\boldsymbol{Q}'_{[f-1,f]}=p_{[f-1,f]}(\boldsymbol{x}_{edge})$
        \STATE $i = f-1$
    \ENDIF
\ENDFOR
\STATE $\boldsymbol{\Phi} = Supersmoother(\mathcal{V}(\boldsymbol{Q}',V))$
\end{algorithmic}
\textbf{Output}: $\boldsymbol{\Phi}\in \mathbb{R}^{V\times F' \times B \times 4}$
\label{alg}
\end{algorithm}

\subsection{Animation Generation and Capturing}
In this stage, we adopt the game engine 3DS Max~\cite{3dsmax} to generate the mesh from the sequence of skeleton representation, to further form the animation. 
The action animations are then showcased using FiveM~\cite{fivem}, a modification platform for GTAV, enabling players to play multi-players on customized dedicated server, in multiple viewpoints. 
We also adopt scene customization, including environmental conditions, character design, and map construction, to craft scenes for these action animations in FiveM. 
This approach accommodates unique action situations and enhanced data diversity.
For further details, please refer to the full paper~\cite{song2024animationbasedaugmentationapproachaction}).


\section{Experimental Results}
\label{sec:results}

\subsection{Representation Evaluation}
To evaluate the efficacy of the 4A pipeline in representing human motion, we execute a series of comparative experiments, comprising both Qualitative and Quantitative Experiments. 
These experiments cover representations of human motion for both major-part (focusing on essential bone joints excluding the hands and feet) and whole-body. 
It is important to note that the major-body representation for evaluation is augmented from the {NTU-RGB+D}~\cite{ntu} dataset, while the whole-body representation is derived from the Human3.6M (H36M)~\cite{h36m} dataset.

\subsubsection{Qualitative Experiment}
\begin{figure*}[ht]
\begin{center}
\centering
\includegraphics[width=\textwidth]{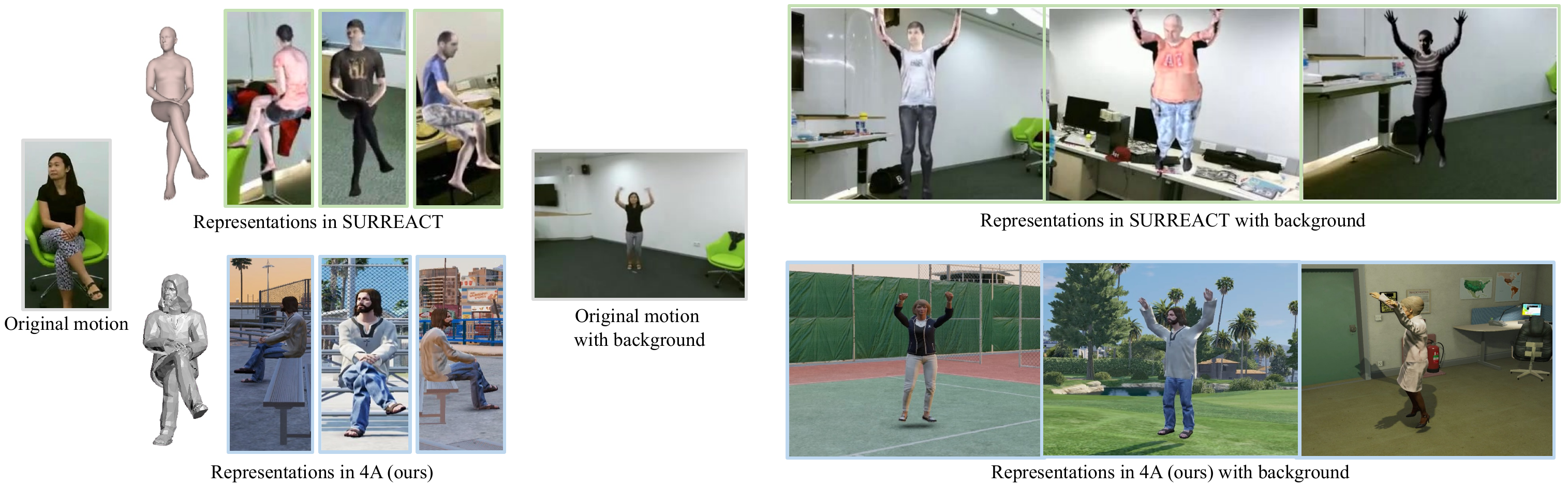}
\end{center}
\vspace{2pt}
\caption{Qualitative results of major-part representation derived from {NTU-RGB+D} dataset, comparing SURREACT with 4A. 
Our pipeline outperforms in terms of fidelity and realism in motion representation and excels in depicting character details. 
Furthermore, it achieves superior integration of characters within their environments, along with enhanced lighting and scene coverage.}
\label{comparison}
\end{figure*}

\begin{figure}[ht]
    \centering
    \includegraphics[width=\linewidth]{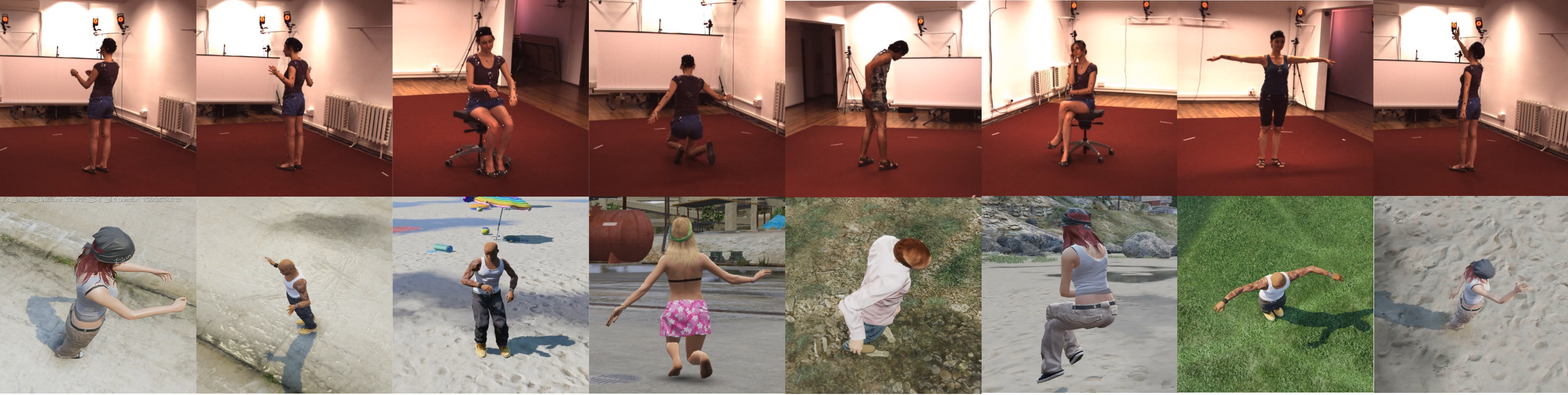}
    \vspace{2pt}
    \caption{Qualitative results of whole-body representation by 4A in multiple viewpoints, comparing with the original RGB frames in H36M. }
    \label{fig:h36m}
\end{figure}

The synthetic representation of human motion derived by {NTU-RGB+D} (comparing with the synthetic representation from SURREACT) and H36M (comparing with the real-world video) are depicted in Figure~\ref{comparison} and~\ref{fig:h36m}.

\subsubsection{Quantitative Experiment}
This evaluation focuses on comparing the effectiveness of models trained on datasets generated by 4A pipeline from discontinuous videos against those trained on datasets generated by other SOTA approaches, or corresponding real-world datasets for human action recognition tasks.

For benchmarking, we establish \textbf{NTU-Original} as a baseline training dataset, incorporating all 49 single-person action classes from the NTU RGB+D dataset, exclusively containing continuous video footage from the real world. 
To mitigate the impact of multiple viewpoints, we selectively use videos where the human is positioned directly in front of the camera (at $0$ degrees).
In contrast, \textbf{NTU-4A}, a dataset created using the 4A process from NTU-Original videos, comprises only synthetic videos derived from discontinuous frames. 
To assess 4A's proficiency in capturing semantic information from discontinuous videos, we employ a \textbf{Random Extracting Strategy} (RES). 
This strategy involves initially extracting one frame from every five frames of each action video, followed by a random extraction of 50\% of these frames, ensuring the avoidance of isolated single frames, to simulate discontinuous video conditions. 
It is noteworthy that, post-RES, the remaining frames constitute only 10\% of the original video frames. 
Comparative training datasets, \textbf{NTU-HMMR} and \textbf{NTU-VIBE}, also generated from NTU-Original utilizing SOTA methods as described in SURREACT~\cite{synthetic_humans}, apply the RES to simulate similar conditions. 
Benchmark results for various action recognition models, evaluated on a real-world video subset of the NTU dataset named \textbf{NTU-Test}, are detailed in Table~\ref{ntu}.

\begin{table*}
\begin{center}
\resizebox{\textwidth}{!}{
\begin{tabular}{c| c c c| c c c|c c c|c c c}
\toprule[2pt]
\multicolumn{1}{c}{\bf{Model Info}} & \multicolumn{3}{c}{\bf{NTU-Original}}& \multicolumn{3}{c}{\bf{NTU-4A (ours)}}& \multicolumn{3}{c}{\bf{NTU-HMMR}}& \multicolumn{3}{c}{\bf{NTU-VIBE}}\\
\toprule[2pt]
\bf{Method}&\bf{Top-1}&\bf{Top-5}&\bf{Mean}&\bf{Top-1}&\bf{Top-5}&\bf{Mean}&\bf{Top-1}&\bf{Top-5}&\bf{Mean}&\bf{Top-1}&\bf{Top-5}&\bf{Mean}\\
\toprule[1pt]
Random &-&-&2.0 &-&-&2.0 &-&-&2.0&-&-&2.0\\
\toprule[0.6pt]
VideoMAE~\cite{videomae} &82.7&86.1&85.7 &82.6&86.4&79.3 &32.6&71.4&30.1 &31.7&70.9&32.3\\
TANet~\cite{model11_tanet} &81.6&87.4&79.5 &77.6&86.2&74.6 &34.5&72.3&32.0 &35.6&73.2&34.5\\
TPN~\cite{model10_tpn} &86.0&90.6&78.9 &80.2&86.5&80.2 &37.1&69.7&39.0 &38.9&71.1&40.3\\
X3D~\cite{x3d} &78.2&85.2&80.9 &68.8&73.3&65.0 &27.9&56.9&32.1 &30.0&59.7&33.1\\
I3D~\cite{dataset05_Kinetics} &76.0&79.7&74.3& 70.1&76.3&70.5 &25.1&60.4&21.5 &26.7&61.0&22.6\\
I3D NL~\cite{model05_NLI3D} &76.9&82.2&76.9& 70.3&77.4&75.4 &30.2&55.7&28.4 &31.1&56.8&29.8\\
\bottomrule[2pt]
\end{tabular}
}
\end{center}
\caption{Benchmark results on NTU-based datasets. NTU-4A, the dataset created by our pipeline from only 10\% of the frames of the NTU-Original dataset, manages to sustain a comparable accuracy when models are trained with real-world continuous videos. 
This performance is notably superior when compared to NTU-HMMR and NTU-VIBE.}
\label{ntu}
\vspace{5pt}
\end{table*}

\begin{table*}[ht]
\begin{center}
\resizebox{\textwidth}{!}{
\begin{tabular}{c |c c c| c c c| c c c| c c c| c c c}
\toprule[2pt]
\multicolumn{1}{c}{\bf{Model Info}} & \multicolumn{3}{c}{\bf{H36M-Original}} & \multicolumn{3}{c}{\bf{H36M-Single}} & \multicolumn{3}{c}{\bf{H36M-Extracted}}
& \multicolumn{3}{c}{\bf{H36M-SingleExtracted}} & \multicolumn{3}{c}{\bf{H36M-4A (Ours)}} \\
\toprule[2pt]
\bf{Method}&\bf{Top-1}&\bf{Top-5}&\bf{Mean}&\bf{Top-1}&\bf{Top-5}&\bf{Mean}&\bf{Top-1}&\bf{Top-5}&\bf{Mean}&\bf{Top-1}&\bf{Top-5}&\bf{Mean}&\bf{Top-1}&\bf{Top-5}&\bf{Mean}\\
\toprule[1pt]
Random &-&-&6.7 &-&-&6.7 &-&-&6.7 &-&-&6.7 &-&-&6.7 \\
\toprule[0.6pt]
VideoMAE~\cite{videomae} &40.2&79.2&37.5 &24.7&77.8&22.5 &15.3&50.7&17.2 &11.4&45.9&12.5 &44.5&88.2&45.0 \\
TANet~\cite{model11_tanet} &33.3&75.0&34.4 &25.5&77.5&19.7 &11.6&49.5&12.4 &13.9&50.1&14.0 &43.9&83.4&46.6 \\
TPN~\cite{model10_tpn} &35.8&78.1&33.6 &22.8&69.4&20.1 &11.4&55.3&11.4 &8.9&48.3&8.9 &51.4&86.7&45.4 \\
X3D~\cite{x3d} &32.1&67.7&29.6 &27.9&75.7&20.3 &11.2&47.3&13.2 &13.3&46.5&12.6 &37.3&80.2&36.5 \\
I3D~\cite{dataset05_Kinetics} &28.7&70.6&30.6 &20.4&69.5&19.5 &11.7&50.0&12.4 &10.5&58.5&9.6 &32.5&75.6&30.6 \\
I3D NL~\cite{model05_NLI3D} &34.2&80.5&32.4 &23.3&68.6&19.6 &11.9&50.5&11.9 &9.8&60.1&11.1 &40.6&78.9&39.7 \\
\bottomrule[2pt]
\end{tabular}
}
\end{center}
\caption{Benchmark results on \textbf{H36M-Original-Test}.
In addition to a notable improvement over H36M-SingleExtracted, the original dataset used for generating H36M-4A, our method enables H36M-4A to achieve results that surpass even those of H36M-Original, considering H36M-4A uses only 2.5\% of the number of frames present in H36M-Original as input.}
\label{tabel:H36M-Original-Test}
\vspace{5pt}
\end{table*}

\begin{table*}[ht]
\begin{center}
\resizebox{\textwidth}{!}{
\begin{tabular}{c| c c c| c c c| c c c| c c c| c c c}
\toprule[2pt]
\multicolumn{1}{c}{\bf{Model Info}} & \multicolumn{3}{c}{\bf{H36M-Original}} & \multicolumn{3}{c}{\bf{H36M-Single}} & \multicolumn{3}{c}{\bf{H36M-Extracted}}
& \multicolumn{3}{c}{\bf{H36M-SingleExtracted}} & \multicolumn{3}{c}{\bf{H36M-4A (Ours)}} \\
\toprule[2pt]
\bf{Method}&\bf{Top-1}&\bf{Top-5}&\bf{Mean}&\bf{Top-1}&\bf{Top-5}&\bf{Mean}&\bf{Top-1}&\bf{Top-5}&\bf{Mean}&\bf{Top-1}&\bf{Top-5}&\bf{Mean}&\bf{Top-1}&\bf{Top-5}&\bf{Mean}\\
\toprule[1pt]
Random &-&-&6.7 &-&-&6.7 &-&-&6.7 &-&-&6.7 &-&-&6.7 \\
\toprule[0.6pt]
VideoMAE~\cite{videomae} &38.3 & 74.3 & 35.0 & 26.1 & 84.3 & 24.5 & 15.9 & 50.9 & 17.4 & 11.2 & 46.7 & 11.7 & 56.1 & 89.7 & 58.2 \\
TANet~\cite{model11_tanet} &30.9 & 81.6 & 35.6 & 24.7 & 70.7 & 18.8 & 11.7 & 52.7 & 13.6 & 14.9 & 45.2 & 14.4 & 51.7 & 75.0 & 51.9 \\
TPN~\cite{model10_tpn} &29.2&80.1&33.6 &16.6&65.4&18.4 &10.6&56.7&11.6 &10.2&51.4&10.4 &65.6&88.1&58.9 \\
X3D~\cite{x3d} &30.5 & 66.9 & 34.1 & 25.3 & 69.2 & 26.0 & 13.4 & 51.6 & 10.4 & 7.5 & 54.5 & 11.3 & 48.6 & 80.3 & 40.0 \\
I3D~\cite{dataset05_Kinetics} &34.2 & 66.4 & 28.9 & 28.4 & 80.7 & 21.0 & 10.1 & 44.4 & 13.2 & 12.1 & 50.8 & 11.6 & 48.8 & 74.0 & 42.4 \\
I3D NL~\cite{model05_NLI3D} &26.4 & 72.2 & 29.3 & 19.2 & 68.0 & 20.9 & 10.8 & 51.1 & 11.7 & 9.7 & 62.4 & 9.7 & 36.6 & 77.2 & 39.5 \\
\bottomrule[2pt]
\end{tabular}
}
\end{center}
\caption{Benchmark results on \textbf{H36M-Segment-Test}.
Apart from the similar improvement observed in testing on H36M-Original-Test, our method enables H36M-4A to attain a better results on segmented video data. This enhancement could be attributed to the automatic segmentation inherent in the Dynamic skeletal interpolation process.}
\label{table:H36M-Segment-Test}
\vspace{5pt}
\end{table*}

Human Whole-body Motion Representation by 4A includes the comprehensive modeling of all 53 bone joints. 
For benchmark purposes, We utilize the Human3.6M~\cite{h36m} (H36M) for comparison in whole-body motion representation. 
The H36M dataset offers two key advantages for augmentation: multiple viewpoints and precise 2D coordinate annotations.
From H36M, we derive four baseline datasets of action recognition task training, as follows:
\begin{enumerate}
\item \textbf{H36M-Original}: This dataset consists of complete action videos from ``S1'', ``S5'', ``S6'', ``S7'' sessions of the H36M dataset, providing four distinct views for each action.
\item \textbf{H36M-Single}: A subset of H36M-Original, while only single view for each action is randomly selected.
\item \textbf{H36M-Extracted}: From H36M-Original, this dataset is formed by adopting RES process to create a discontinuous dataset.
\item \textbf{H36M-SingleExtracted}: Applying RES process as H36M-Extracted, but starting from the H36M-Single dataset, to create a version that is both single-view and sparsely sampled.
\end{enumerate}
Additionally, for the purpose of testing, two specific datasets derived from different sections of the H36M dataset were used:
\begin{enumerate}
\item \textbf{H36M-Original-Test}: This test dataset includes all videos from sections ``S8'', ``S9'', ``S11'' of H36M, featuring each action captured from four distinct views.
\item \textbf{H36M-Segment-Test}:  In this dataset, every video from H36M-Original-Test has been manually segmented into individual unit actions, with the removal of indistinguishable clips.
\end{enumerate}

For our experimental purposes, we develop the \textbf{H36M-4A} dataset for training. 
This dataset is derived from each frame of the H36M-SingleExtracted dataset, utilizing the 4A framework for generating whole-body representations. 
The primary objective is to assess 4A's capabilities of multi-viewpoint data generation and frame interpolation.
The benchmark results on various action recognition methods, evaluated using H36M-Original-Test and H36M-Segment-Test as test datasets, are documented in Table~\ref{tabel:H36M-Original-Test} and Table~\ref{table:H36M-Segment-Test} respectively.

\subsection{Component-wise Comparative Analysis}
\subsubsection{Q-GCN}
We initiate our evaluation by performing comparative experiments on the task of lifting 2D coordinates to 3D orientations. 
Due to the limited exploration in this area~\cite{quaternet}, we employ our mean Average Angular Distance (mAAD) loss for evaluation. 
All baseline methods included in the comparison are specifically tailored for the 2D to 3D pose lifting task. 
The models are both trained and evaluated on the {H3WB}~\cite{h3wb} dataset. 
The results of these comparisons are detailed in Table~\ref{tabel:qgcn}.

\begin{table}[h]
\begin{center}
\resizebox{0.5\textwidth}{!}{
\begin{tabular}{c c c c c c}
\toprule[2pt]
\bf{Method} & \bf{Whole-body} & \bf{Major-part} & \bf{Upper-body} & \bf{Lower-body} &\bf{Hands}\\
\toprule[1pt]
SMPL-X~\cite{shape_SMPLx} &     123 & 72 & 89 & 64 & 167\\
Jointformer~\cite{jointformer} & 77 & 66 & 72 & 49 & 103\\
GLA-GCN~\cite{pe_glagcn} &     79 & 54 & 63 & 41 & 91\\
Q-GCN (ours) & 67 & 32 & 41 & 27 & 83\\
\bottomrule[2pt]
\end{tabular}
}
\end{center}
\caption{Comparative results of different methods 2D to 3D orientation lifting task. Results are presented in terms of the mean Average Angular Distance (mAAD) loss in table, scaled by $10^{3}$, with a lower score indicating superior performance. 
Our method Q-GCN, outperforms all other methods, demonstrating the highest effectiveness in this task. }
\label{tabel:qgcn}
\vspace{5pt}
\end{table}

\subsubsection{DSI}
\begin{figure*}[ht]
    \centering
    \includegraphics[width=\linewidth]{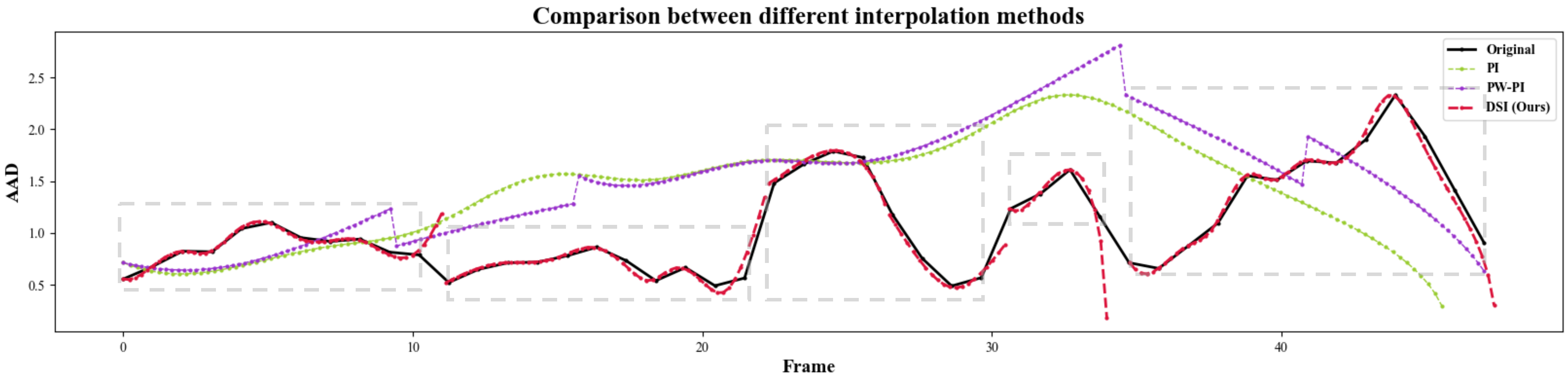}
    \vspace{2pt}
    \caption{Comparative analysis of different interpolation method.
In the provided figure, the plots represented in black, green, purple, and red correspond to the Absolute Angular Distance (AAD) of the original sequence (Original), and sequences interpolated using Polynomial Interpolation (PI), Point-wise Polynomial Interpolation (PW-PI), and Dynamic Skeletal Interpolation (DSI), respectively. The five gray dot boxes illustrate the five randomly selected Quaternion sequence segments.
The PI method yields a smooth sequence but lacks dynamic segmentation capabilities. 
PW-PI produces a segmented sequence but leads to a decrease in movement amplitude during interpolation, evident from the overly smoothed curve.
DSI stands out by not only accurately segmenting the sequence but also preserving the semantic integrity of the motion, showcasing its superior capability in maintaining both fluidity and semantic richness in the interpolated sequence.}
    \label{fig:DSI}
\end{figure*}

In this section, we assess the performance of our Dynamic Skeletal Interpolation (DSI) compared to other interpolation methods. 
We introduce the Absolute Angular Distance (AAD) metric to measure the angular distance from the current position to the initial position, with the initial position set along the $x$-axis where the Quaternion equals 1. 
This metric allows us to track the fluidity of motion across a sequence of frames. 
Furthermore, to evaluate DSI's efficacy in segmenting Quaternion sequences, we employ a strategy similar to the Random Extracting Strategy (RES). 
This involves extracting one frame from every five in a continuous Quaternion sequence, then randomly selecting five segments of varying lengths to compile into a complete sequence (termed the Original sequence). 
We compare DSI against Polynomial Interpolation (PI) and Point-wise Polynomial Interpolation (PW-PI). 
The interpolation results from these three algorithms, alongside the Original sequence, are illustrated in Figure~\ref{fig:DSI}.

\subsection{Evaluation on In-the-Wild Video}
In this part, we extend the evaluation of our approach to in-the-wild videos. 
In-the-wild videos are captured in uncontrolled and natural environments, presenting a greater challenge for processing and analysis. 
We employ the same evaluation strategy as described in~\cite{synthetic_humans}.
Initially, we utilize the ResNeXt-101~\cite{ResNeXt-101} 3D CNN model, pre-trained on the Mini-Kinetics-200~\cite{k200} dataset, as our feature extractor. 
Subsequently, we employ the Kinetics-15, a 15-class subset of the Kinetics-400 dataset, devised by~\cite{synthetic_humans}. 
From these 15 actions, one training video per class is randomly selected for synthetic data generation, with the remaining 725 videos designated for validation purposes. 
Additionally, random selection and nearest neighbor approaches, leveraging pre-trained features, serve as baselines for comparison. 
The results provided by~\cite{synthetic_humans} are included for comparison. 
The outcomes of various methods are detailed in Table~\ref{tabel:In-the-Wild}.

\begin{table}[h]
\begin{center}
{
\begin{tabular}{c | c c c}
\toprule[2pt]
& \multicolumn{3}{c}{\bf{Accuracy (\%)}} \\
\bf{Method}&{RGB}&{Flow}&{RGB+Flow}\\
\toprule[1pt]
Random &6.7 & 6.7 &6.7 \\
Nearest neighbor & 8.6 & 13.1 & 13.9\\
\toprule[0.5pt]
Real & 26.2 &20.6 &28.4\\
SURREACT & 9.4 & 10.3 & 11.6\\
Synth + Real & 32.7& 22.3 &34.6\\
\toprule[0.5pt]
4A &  11.1 & 10.8 & 12.4\\
4A + Real & 36.5 & 24.2 & 37.1\\
\bottomrule[2pt]
\end{tabular}
}
\end{center}
\caption{Evaluation on In-the-Wild Videos.
It reveals that training solely with synthetic data, regardless of the approach used, results in significantly lower performance compared to training with real data. 
Nonetheless, our method achieves a slight improvement when combined with real training data, achieving higher accuracy than training exclusively with real data or using the SURREACT. }
\label{tabel:In-the-Wild}
\vspace{5pt}
\end{table}


\section{Conclusion}
\label{sec:conclusion}
In this paper, we present a comprehensive study on enhancing action recognition models using synthetic data augmentation, particularly focusing on the challenges posed by discontinuous frames and the limitations of existing methods. 
We propose 4A pipeline, leveraging game engine technology to generate sophisticated semantic representations of human motion. 
Our comparative analyses across various benchmarks demonstrate the superior performance of 4A pipeline in maintaining high accuracy levels even when trained on significantly reduced data. 
The components (Q-GCN and DSI) of our pipeline are pivotal in achieving these results, effectively capturing the nuanced dynamics of human motion and facilitating the generation of semantically rich synthetic datasets.
Evaluation on in-the-wild videos further validates the effectiveness of our approach, illustrating a enhanced ability of understanding actions in the real world.

\newpage
\bibliography{ref}

\begin{thebibliography}{56}
\providecommand{\natexlab}[1]{#1}
\providecommand{\url}[1]{\texttt{#1}}
\expandafter\ifx\csname urlstyle\endcsname\relax
  \providecommand{\doi}[1]{doi: #1}\else
  \providecommand{\doi}{doi: \begingroup \urlstyle{rm}\Url}\fi

\bibitem[Ardianto and Hang(2019)]{gamedataset04_GTA360}
S.~Ardianto and H.-M. Hang.
\newblock Nctu-gtav360: A 360° action recognition video dataset.
\newblock In \emph{2019 IEEE 21st International Workshop on Multimedia Signal Processing (MMSP)}, pages 1--5, 2019.

\bibitem[{Autodesk}(2023)]{3dsmax}
{Autodesk}.
\newblock 3ds max 2023.
\newblock \url{https://www.autodesk.co.jp/products/3ds-max}, 2023.

\bibitem[Bloom et~al.(2012)Bloom, Makris, and Argyriou]{gamedataset01_G3D}
V.~Bloom, D.~Makris, and V.~Argyriou.
\newblock G3d: A gaming action dataset and real time action recognition evaluation framework.
\newblock In \emph{2012 IEEE Computer society conference on computer vision and pattern recognition workshops}, pages 7--12. IEEE, 2012.

\bibitem[Cao et~al.(2020)Cao, Gao, Mangalam, Cai, Vo, and Malik]{gamedataset02_GTAIM}
Z.~Cao, H.~Gao, K.~Mangalam, Q.~Cai, M.~Vo, and J.~Malik.
\newblock Long-term human motion prediction with scene context.
\newblock \emph{CoRR}, abs/2007.03672, 2020.
\newblock URL \url{https://arxiv.org/abs/2007.03672}.

\bibitem[Carreira and Zisserman(2017)]{dataset05_Kinetics}
J.~Carreira and A.~Zisserman.
\newblock Quo vadis, action recognition? {A} new model and the kinetics dataset.
\newblock \emph{CoRR}, abs/1705.07750, 2017.
\newblock URL \url{http://arxiv.org/abs/1705.07750}.

\bibitem[de~Souza12 et~al.(2017)de~Souza12, Gaidon, Cabon, and L{\'o}pez]{relate_ar_1}
C.~R. de~Souza12, A.~Gaidon, Y.~Cabon, and A.~M. L{\'o}pez.
\newblock Procedural generation of videos to train deep action recognition networks.
\newblock \emph{CVPR}, 2017.

\bibitem[{dexyfex}(2023)]{codewroker}
{dexyfex}.
\newblock Codeworker.
\newblock \url{https://de.gta5-mods.com/tools/codewalker-gtav-interactive-3d-map}, 2023.
\newblock Accessed March 8, 2023.

\bibitem[Diba et~al.(2019)Diba, Fayyaz, Sharma, Paluri, Gall, Stiefelhagen, and Gool]{dataset01_HVU}
A.~Diba, M.~Fayyaz, V.~Sharma, M.~Paluri, J.~Gall, R.~Stiefelhagen, and L.~V. Gool.
\newblock Holistic large scale video understanding.
\newblock \emph{CoRR}, abs/1904.11451, 2019.
\newblock URL \url{http://arxiv.org/abs/1904.11451}.

\bibitem[Fabbri et~al.(2018)Fabbri, Lanzi, Calderara, Palazzi, Vezzani, and Cucchiara]{jta}
M.~Fabbri, F.~Lanzi, S.~Calderara, A.~Palazzi, R.~Vezzani, and R.~Cucchiara.
\newblock Learning to detect and track visible and occluded body joints in a virtual world.
\newblock In \emph{ECCV}, 2018.

\bibitem[Feichtenhofer(2020)]{x3d}
C.~Feichtenhofer.
\newblock X3d: Expanding architectures for efficient video recognition, 2020.

\bibitem[{Fivem}(2023)]{fivem}
{Fivem}.
\newblock Fivem.
\newblock \url{https://fivem.net}, 2023.
\newblock Accessed March 8, 2023.

\bibitem[Friedman(1984)]{supersmoother}
J.~H. Friedman.
\newblock \emph{A variable span smoother}.
\newblock Laboratory for Computational Statistics, Department of Statistics, Stanford~…, 1984.

\bibitem[Ghezelghieh et~al.(2016)Ghezelghieh, Kasturi, and Sarkar]{relate_sysshape_2}
M.~F. Ghezelghieh, R.~Kasturi, and S.~Sarkar.
\newblock Learning camera viewpoint using cnn to improve 3d body pose estimation.
\newblock In \emph{3DV}, pages 685--693. IEEE, 2016.

\bibitem[Hara et~al.(2017)Hara, Kataoka, and Satoh]{ResNeXt-101}
K.~Hara, H.~Kataoka, and Y.~Satoh.
\newblock Can spatiotemporal 3d cnns retrace the history of 2d cnns and imagenet?
\newblock \emph{CoRR}, abs/1711.09577, 2017.
\newblock URL \url{http://arxiv.org/abs/1711.09577}.

\bibitem[He et~al.(2015)He, Zhang, Ren, and Sun]{resnet}
K.~He, X.~Zhang, S.~Ren, and J.~Sun.
\newblock Deep residual learning for image recognition.
\newblock \emph{CoRR}, abs/1512.03385, 2015.
\newblock URL \url{http://arxiv.org/abs/1512.03385}.

\bibitem[Hu et~al.(2021)Hu, Zhang, Zhan, Zhang, and Wong]{CDGCN}
W.~Hu, C.~Zhang, F.~Zhan, L.~Zhang, and T.~Wong.
\newblock Conditional directed graph convolution for 3d human pose estimation.
\newblock \emph{CoRR}, abs/2107.07797, 2021.
\newblock URL \url{https://arxiv.org/abs/2107.07797}.

\bibitem[Ionescu et~al.(2014)Ionescu, Papava, Olaru, and Sminchisescu]{h36m}
C.~Ionescu, D.~Papava, V.~Olaru, and C.~Sminchisescu.
\newblock Human3.6m: Large scale datasets and predictive methods for 3d human sensing in natural environments.
\newblock \emph{IEEE Transactions on Pattern Analysis and Machine Intelligence}, 2014.

\bibitem[Jin et~al.(2020)Jin, Xu, Xu, Wang, Liu, Qian, Ouyang, and Luo]{coco_whole-body}
S.~Jin, L.~Xu, J.~Xu, C.~Wang, W.~Liu, C.~Qian, W.~Ouyang, and P.~Luo.
\newblock Whole-body human pose estimation in the wild.
\newblock In \emph{Proceedings of the ECCV}, 2020.

\bibitem[Kanazawa et~al.(2018)Kanazawa, Zhang, Felsen, and Malik]{3d_rec_hmmr}
A.~Kanazawa, J.~Y. Zhang, P.~Felsen, and J.~Malik.
\newblock Learning 3d human dynamics from video.
\newblock \emph{CoRR}, abs/1812.01601, 2018.

\bibitem[Kipf and Welling(2016)]{gcn}
T.~N. Kipf and M.~Welling.
\newblock Semi-supervised classification with graph convolutional networks.
\newblock \emph{CoRR}, abs/1609.02907, 2016.

\bibitem[Korban and Li(2020)]{ar_ddgcn}
M.~Korban and X.~Li.
\newblock Ddgcn: A dynamic directed graph convolutional network for action recognition.
\newblock In \emph{ECCV}, pages 761--776. Springer, 2020.

\bibitem[Li et~al.(2024)Li, Song, Chen, and Demachi]{LI2024121367}
Z.~Li, X.~Song, S.~Chen, and K.~Demachi.
\newblock Data, language and graph-based reasoning methods for identification of human malicious behaviors in nuclear security.
\newblock \emph{Expert Systems with Applications}, 236:\penalty0 121367, 2024.
\newblock ISSN 0957-4174.

\bibitem[Liu and Mian(2017)]{relate_ar_3}
J.~Liu and A.~Mian.
\newblock Learning human pose models from synthesized data for robust {RGB-D} action recognition.
\newblock \emph{CoRR}, abs/1707.00823, 2017.
\newblock URL \url{http://arxiv.org/abs/1707.00823}.

\bibitem[Liu et~al.(2020{\natexlab{a}})Liu, Shen, Wang, Chen, Cheung, and Asari]{RN013}
R.~Liu, J.~Shen, H.~Wang, C.~Chen, S.-c. Cheung, and V.~Asari.
\newblock Attention mechanism exploits temporal contexts: Real-time 3d human pose reconstruction.
\newblock In \emph{CVPR}, pages 5064--5073, 2020{\natexlab{a}}.

\bibitem[Liu et~al.(2020{\natexlab{b}})Liu, Wang, Wu, Qian, and Lu]{model11_tanet}
Z.~Liu, L.~Wang, W.~Wu, C.~Qian, and T.~Lu.
\newblock Tam: Temporal adaptive module for video recognition.
\newblock \emph{arXiv preprint arXiv:2005.06803}, 2020{\natexlab{b}}.

\bibitem[Loper et~al.(2015)Loper, Mahmood, Romero, Pons-Moll, and Black]{shape_SMPLx}
M.~Loper, N.~Mahmood, J.~Romero, G.~Pons-Moll, and M.~J. Black.
\newblock {SMPL}: A skinned multi-person linear model.
\newblock \emph{ACM Transactions on Graphics, (Proc. SIGGRAPH Asia)}, 34\penalty0 (6):\penalty0 248:1--248:16, Oct. 2015.

\bibitem[Lutz et~al.(2022)Lutz, Blythman, Ghostal, Matthew, Simms, and Smolic]{jointformer}
S.~Lutz, R.~Blythman, K.~Ghostal, M.~Matthew, C.~Simms, and A.~Smolic.
\newblock Jointformer: Single-frame lifting transformer with error prediction and refinement for 3d human pose estimation.
\newblock \emph{ICPR}, 2022.

\bibitem[Lv and Nevatia(2007)]{relate_ar_4}
F.~Lv and R.~Nevatia.
\newblock Single view human action recognition using key pose matching and viterbi path searching.
\newblock In \emph{CVPR}, pages 1--8, 2007.
\newblock \doi{10.1109/CVPR.2007.383131}.

\bibitem[Ma et~al.(2021)Ma, Su, Wang, Ci, and Wang]{ma2021context}
X.~Ma, J.~Su, C.~Wang, H.~Ci, and Y.~Wang.
\newblock Context modeling in 3d human pose estimation: A unified perspective.
\newblock In \emph{CVPR}, pages 6238--6247, 2021.

\bibitem[Osman et~al.(2020)Osman, Bolkart, and Black]{shape_STAR:2020}
A.~A.~A. Osman, T.~Bolkart, and M.~J. Black.
\newblock {STAR}: A sparse trained articulated human body regressor.
\newblock In \emph{ECCV}, pages 598--613, 2020.
\newblock URL \url{https://star.is.tue.mpg.de}.

\bibitem[Pavllo et~al.(2018{\natexlab{a}})Pavllo, Feichtenhofer, Grangier, and Auli]{3D_PE_semi}
D.~Pavllo, C.~Feichtenhofer, D.~Grangier, and M.~Auli.
\newblock 3d human pose estimation in video with temporal convolutions and semi-supervised training.
\newblock \emph{CoRR}, abs/1811.11742, 2018{\natexlab{a}}.

\bibitem[Pavllo et~al.(2018{\natexlab{b}})Pavllo, Grangier, and Auli]{quaternet}
D.~Pavllo, D.~Grangier, and M.~Auli.
\newblock Quaternet: {A} quaternion-based recurrent model for human motion.
\newblock \emph{CoRR}, abs/1805.06485, 2018{\natexlab{b}}.

\bibitem[Pham et~al.(2022)Pham, Khoudour, Crouzil, Zegers, and Velastin]{semantic1}
H.~H. Pham, L.~Khoudour, A.~Crouzil, P.~Zegers, and S.~A. Velastin.
\newblock Video-based human action recognition using deep learning: A review, 2022.

\bibitem[Qian et~al.(2018)Qian, Fu, Xiang, Wang, Qiu, Wu, Jiang, and Xue]{relate_reid}
X.~Qian, Y.~Fu, T.~Xiang, W.~Wang, J.~Qiu, Y.~Wu, Y.-G. Jiang, and X.~Xue.
\newblock Pose-normalized image generation for person re-identification.
\newblock In \emph{Proceedings of the ECCV}, pages 650--667, 2018.

\bibitem[Rahmani and Mian(2015)]{relate_ar_5}
H.~Rahmani and A.~Mian.
\newblock Learning a non-linear knowledge transfer model for cross-view action recognition.
\newblock In \emph{CVPR}, pages 2458--2466, 2015.
\newblock \doi{10.1109/CVPR.2015.7298860}.

\bibitem[Roitberg et~al.(2021)Roitberg, Schneider, Djamal, Seibold, Rei{\ss}, and Stiefelhagen]{gamedataset03_SIM}
A.~Roitberg, D.~Schneider, A.~Djamal, C.~Seibold, S.~Rei{\ss}, and R.~Stiefelhagen.
\newblock Let's play for action: Recognizing activities of daily living by learning from life simulation video games.
\newblock \emph{CoRR}, abs/2107.05617, 2021.
\newblock URL \url{https://arxiv.org/abs/2107.05617}.

\bibitem[Romero et~al.(2017)Romero, Tzionas, and Black]{shape_MANO:SIGGRAPHASIA:2017}
J.~Romero, D.~Tzionas, and M.~J. Black.
\newblock Embodied hands: Modeling and capturing hands and bodies together.
\newblock \emph{ACM Transactions on Graphics, (Proc. SIGGRAPH Asia)}, 36\penalty0 (6), Nov. 2017.

\bibitem[Seo et~al.(2015)Seo, Han, Lee, and Kim]{intro_safe}
J.~Seo, S.~Han, S.~Lee, and H.~Kim.
\newblock Computer vision techniques for construction safety and health monitoring.
\newblock \emph{Advanced Engineering Informatics}, 29\penalty0 (2):\penalty0 239--251, 2015.
\newblock ISSN 1474-0346.

\bibitem[Shahroudy et~al.(2016)Shahroudy, Liu, Ng, and Wang]{ntu}
A.~Shahroudy, J.~Liu, T.-T. Ng, and G.~Wang.
\newblock Ntu rgb+d: A large scale dataset for 3d human activity analysis.
\newblock In \emph{CVPR}, pages 1010--1019, 2016.

\bibitem[Shi et~al.(2019)Shi, Zhang, Cheng, and Lu]{DGCN}
L.~Shi, Y.~Zhang, J.~Cheng, and H.~Lu.
\newblock Skeleton-based action recognition with directed graph neural networks.
\newblock In \emph{CVPR}, pages 7912--7921, 2019.

\bibitem[Shotton et~al.(2011)Shotton, Fitzgibbon, Cook, Sharp, Finocchio, Moore, Kipman, and Blake]{relate_poseseg_1}
J.~Shotton, A.~Fitzgibbon, M.~Cook, T.~Sharp, M.~Finocchio, R.~Moore, A.~Kipman, and A.~Blake.
\newblock Real-time human pose recognition in parts from single depth images.
\newblock In \emph{CVPR 2011}, pages 1297--1304. Ieee, 2011.

\bibitem[Song et~al.(2024)Song, Li, Chen, Cai, and Demachi]{song2024animationbasedaugmentationapproachaction}
X.~Song, Z.~Li, S.~Chen, X.-Q. Cai, and K.~Demachi.
\newblock An animation-based augmentation approach for action recognition from discontinuous video, 2024.
\newblock URL \url{https://arxiv.org/abs/2404.06741}.

\bibitem[Su et~al.(2015)Su, Qi, Li, and Guibas]{render4cnn}
H.~Su, C.~R. Qi, Y.~Li, and L.~J. Guibas.
\newblock Render for {CNN:} viewpoint estimation in images using cnns trained with rendered 3d model views.
\newblock \emph{CoRR}, abs/1505.05641, 2015.
\newblock URL \url{http://arxiv.org/abs/1505.05641}.

\bibitem[Sun et~al.(2019)Sun, Xiao, Liu, and Wang]{hrnet}
K.~Sun, B.~Xiao, D.~Liu, and J.~Wang.
\newblock Deep high-resolution representation learning for human pose estimation.
\newblock In \emph{CVPR (CVPR)}, June 2019.

\bibitem[Tong et~al.(2022)Tong, Song, Wang, and Wang]{videomae}
Z.~Tong, Y.~Song, J.~Wang, and L.~Wang.
\newblock Video{MAE}: Masked autoencoders are data-efficient learners for self-supervised video pre-training.
\newblock In \emph{Advances in Neural Information Processing Systems}, 2022.

\bibitem[Varol et~al.(2017)Varol, Romero, Martin, Mahmood, Black, Laptev, and Schmid]{surreal}
G.~Varol, J.~Romero, X.~Martin, N.~Mahmood, M.~J. Black, I.~Laptev, and C.~Schmid.
\newblock Learning from synthetic humans.
\newblock \emph{CoRR}, abs/1701.01370, 2017.
\newblock URL \url{http://arxiv.org/abs/1701.01370}.

\bibitem[Varol et~al.(2021)Varol, Laptev, Schmid, and Zisserman]{synthetic_humans}
G.~Varol, I.~Laptev, C.~Schmid, and A.~Zisserman.
\newblock Synthetic humans for action recognition from unseen viewpoints.
\newblock In \emph{IJCV}, 2021.

\bibitem[Wang et~al.(2016)Wang, Xiong, Wang, Qiao, Lin, Tang, and Van~Gool]{model01_TSN}
L.~Wang, Y.~Xiong, Z.~Wang, Y.~Qiao, D.~Lin, X.~Tang, and L.~Van~Gool.
\newblock Temporal segment networks: Towards good practices for deep action recognition.
\newblock In \emph{ECCV}, pages 20--36. Springer, 2016.

\bibitem[Wang et~al.(2018)Wang, Girshick, Gupta, and He]{model05_NLI3D}
X.~Wang, R.~Girshick, A.~Gupta, and K.~He.
\newblock Non-local neural networks.
\newblock \emph{CVPR}, 2018.

\bibitem[Xie et~al.(2017)Xie, Sun, Huang, Tu, and Murphy]{k200}
S.~Xie, C.~Sun, J.~Huang, Z.~Tu, and K.~Murphy.
\newblock Rethinking spatiotemporal feature learning for video understanding.
\newblock \emph{CoRR}, abs/1712.04851, 2017.
\newblock URL \url{http://arxiv.org/abs/1712.04851}.

\bibitem[Yan et~al.(2018)Yan, Xiong, and Lin]{model12_stgcn}
S.~Yan, Y.~Xiong, and D.~Lin.
\newblock Spatial temporal graph convolutional networks for skeleton-based action recognition.
\newblock In \emph{AAAI}, 2018.

\bibitem[Yang et~al.(2020)Yang, Xu, Shi, Dai, and Zhou]{model10_tpn}
C.~Yang, Y.~Xu, J.~Shi, B.~Dai, and B.~Zhou.
\newblock Temporal pyramid network for action recognition.
\newblock In \emph{CVPR (CVPR)}, 2020.

\bibitem[Yu et~al.(2023)Yu, Zhang, Liu, Zhong, Liu, and Chen]{pe_glagcn}
B.~X. Yu, Z.~Zhang, Y.~Liu, S.-h. Zhong, Y.~Liu, and C.~W. Chen.
\newblock Gla-gcn: Global-local adaptive graph convolutional network for 3d human.
\newblock \emph{arXiv preprint arXiv:2307.05853}, 2023.

\bibitem[Zhao et~al.(2019)Zhao, Peng, Tian, Kapadia, and Metaxas]{pr_semgcn}
L.~Zhao, X.~Peng, Y.~Tian, M.~Kapadia, and D.~N. Metaxas.
\newblock Semantic graph convolutional networks for 3d human pose regression.
\newblock In \emph{CVPR}, pages 3425--3435, 2019.

\bibitem[Zhou et~al.(2019)Zhou, Han, Jiang, Jia, and Lu]{RN021}
K.~Zhou, X.~Han, N.~Jiang, K.~Jia, and J.~Lu.
\newblock Hemlets pose: Learning part-centric heatmap triplets for accurate 3d human pose estimation.
\newblock In \emph{ICCV}, pages 2344--2353, 2019.

\bibitem[Zhu et~al.(2023)Zhu, Samet, and Picard]{h3wb}
Y.~Zhu, N.~Samet, and D.~Picard.
\newblock H3wb: Human3.6m 3d wholebody dataset and benchmark.
\newblock In \emph{ICCV}, pages 20166--20177, October 2023.

\end{thebibliography}
\clearpage

\setcounter{page}{1}
\setcounter{section}{0}

\newpage
\twocolumn[
\centering
\Large
\vspace{0.5em}\textbf{**Appendix**}\\
\vspace{1.0em}
] 

\renewcommand\thesection{\Alph{section}}
\section{Extended Introduction to 4A}

This section offers supplementary details to further elucidate the 4A introduction.

\subsection{Interpretation of Terminologies}
In 4A, we define several key terms to facilitate the creation and modification of action recognition datasets:
\begin{itemize}
    \item \textbf{Character}: Refers to the main player executing the action, captured by the camera.
    \item \textbf{Scene}: Denotes the environment where the action takes place. It includes natural game elements like weather and time of day, character conditions such as appearance and behavior style, and the scenery within the camera's view, such as character location and surroundings.
    \item \textbf{Animation}: Represents the sequence of movements or actions performed by the character.
    \item \textbf{Entity}: Encompasses all in-game objects, ranging from large structures like buildings to small items like books. Entities possess attributes beyond appearance, including weight, texture, collision volume, etc. 
    \item \textbf{Ped (Pedestrian)}: In GTAV, this term refers to non-player characters or pedestrian models within the game. FiveM allows players to customize their in-game pedestrian models. 
\end{itemize}

\subsection{Theoretical Underpinnings}
This section provides additional explanations of the concepts introduced in the main body of the paper.

\subsubsection{Skeleton Configuration}
\begin{figure}[ht]
    \centering
    \includegraphics[width=0.6\linewidth]{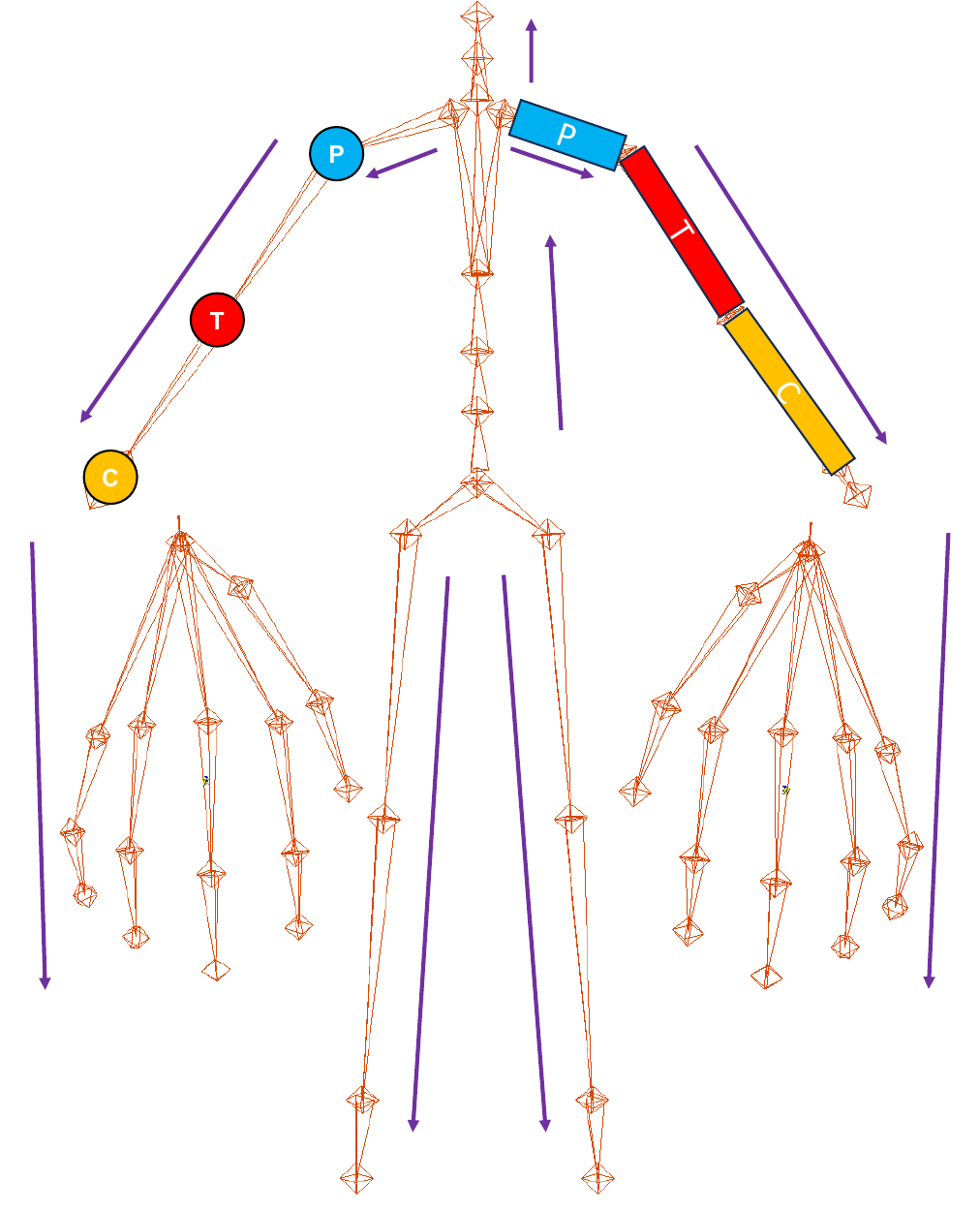}
    \caption{Configuration of whole-body skeleton graph: Red, yellow and blue node (bone) joint stand for target, child and parent node (bone) joint. Violet arrows denote the hierarchical inheritance directions from parent joint to child joint. The whole structure contains 54 node joints and 53 bone joints. }
    \label{fig:graph_config}
\end{figure}

The skeleton configuration is shown in Figure~\ref{fig:graph_config}.

\subsubsection{Q-GCN}
We use H3WB~\cite{h3wb} to derive the training dataset for Q-GCN. 
We employ coordinate transformation to convert the 3D coordinates of discrete joint nodes within the human body skeleton into the rotations of each bone joint in. 
Additionally, we articulate these bone joints using 3D vectors.
For example, the vector representing the upper arm is derived by subtracting the coordinates of the shoulder node from those of the elbow node: 
\begin{equation}
\Vec{v}=\boldsymbol{p}_{elbow}-\boldsymbol{p}_{shoulder}
\end{equation}

\begin{figure}[ht]
\centering
\includegraphics[width=0.8\columnwidth]{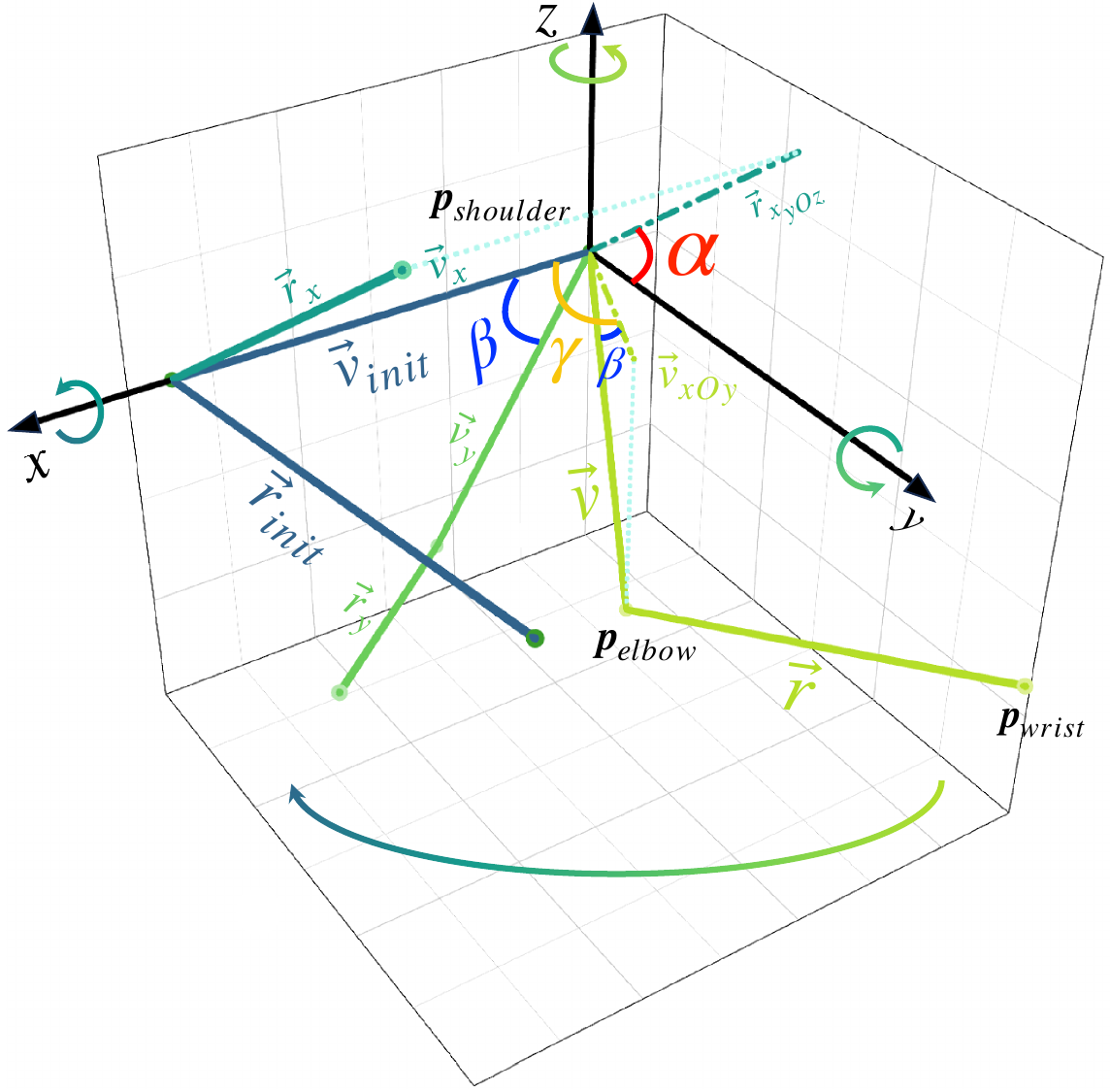} 
\caption{Schematic diagram of Euler angle calculation:
$\boldsymbol{p}_{shoulder}$ (rotation center), $\boldsymbol{p}_{elbow}$, and $\boldsymbol{p}_{wrist}$ represent the nodes corresponding to the shoulder, elbow and wrist joints in 3D World Coordinate System respectively. 
Given the initial target vector $\Vec{v}_{init}$ and initial reference vector $\Vec{r}_{init}$, and proceeding to rotate them sequentially around the $x,y \text{ and } z$ axes by angles $\alpha,\beta, \text{ and } \gamma$ respectively, $(\Vec{v}_x,\Vec{r}_x)$, $(\Vec{v}_y,\Vec{r}_y)$ and finally $(\Vec{v},\Vec{r})$ can be obtained. 
The diagram displays a gradient arrow, signifying the Euler angle calculation order, which is in reverse to the rotation sequence.
Consequently, we calculate $\gamma$ as the rotation angle from the positive $x$-axis to $\vec{v}_{xOy}$, which representing the projection of vector $\vec{v}$ onto the $xOy$ plane. 
Likewise, $\beta$ is inferred as the rotation angle between vector $\vec{v}$ and the $xOy$ plane. 
Lastly, $\alpha$ is calculated as the angle spanning from the positive $y$-axis to $\vec{r}_{x_{yOz}}$, the projection of $\vec{r}_{x}$ onto the $yOz$ plane.}
\label{euler_angle}
\end{figure}

Therefore, poses can be articulated by the rotations of each bone vector from the initial position to the current position. 
Initially, We compute the static Euler angles of each vector in World Coordinate System in $x-y-z$ sequence, utilizing the corresponding coordinates of skeleton nodes. 
To illustrate, Figure~\ref{euler_angle} demonstrates the calculation of the Euler angles $(\alpha,\beta,\gamma)$ for the upper arm vector $\Vec{v}$, with shoulder joint node $\boldsymbol{p}_{shoulder}$ as the center of rotation. 
Hence, Euler angles $\gamma$ and $\beta$ can be calculated as:
\begin{equation}
    \vec{v}_{xOy}=\vec{v}-\langle \vec{v},\vec{n}_{z} \rangle 
\end{equation}
\begin{equation}
    \gamma=\arccos{
    \frac{\vec{v}_{xOy} \cdot \vec{i}_{x}}
    {||\vec{v}_{xOy}||}    
}
\end{equation}
\begin{equation}
    \beta=\arccos{
    \frac{\vec{v}_{xOy} \cdot \vec{v}}
    {||\vec{v}_{xOy}|| \cdot ||\vec{v}||}   
    }
\end{equation}
where $\vec{n}_{z}$ represents the normal vector of $xOy$ plane and $\vec{i}_{x}$ signifies the unit vector along $x$-axis. 

Nonetheless, when a 3D vector is parallel to the $x$-axis, it becomes incapable of representing a rotation about the $x$-axis.
To address this, we introduce a reference vector $\vec{r}$, represented by the vector formed by the child bone joint of the target bone joint, such as the forearm (child) and upper arm (parent), aiding in the calculation of $\alpha$. 
Consequently, $\alpha$ can be calculated using $\vec{r}$, $\gamma$, and $\beta$ as follows:

\begin{equation}
    \vec{r}_{x}=RM^{-1}_{z}(\gamma) \cdot RM^{-1}_{y}(\beta) \cdot \vec{r}
\end{equation}
\begin{equation}
    \alpha=\arccos{
    \frac{(\vec{r}_{x}-\langle \vec{r}_{x},\vec{n}_{x} \rangle ) \cdot \vec{i}_{y}}
    {||\vec{r}_{x}-\langle \vec{r}_{x},\vec{n}_{x} \rangle ||}    
}
\end{equation}
where $\vec{r}_{x}$ represents the vector $\vec{r}$ rotated around $x$-axis by angle $\alpha$. $RM^{-1}_{z}(\gamma)$ and $RM^{-1}_{y}(\beta)$ denote the inverse rotation metrics for angles $\gamma$ and $\beta$ respectively. $\vec{n}_{x}$ denotes the normal vector to $yOz$ plane, and $\vec{i}_{y}$ is the unit vector along $y$-axis.

However, to avoid the potential Gimbal Lock issue associated with Euler angles, we convert these angles into Quaternions\footnotemark[1] for rotation representation. 
The transformation from Euler angles to Quaternions is illustrated below:
\begin{equation}
    \boldsymbol{q}=q_{x}\boldsymbol{i}+q_{y}\boldsymbol{j}+q_{z}\boldsymbol{k}+q_{w}
\end{equation}
\begin{equation}
    q_{x}=\sin{(\theta/2)}\cdot \cos{\alpha} 
\end{equation}
\begin{equation}
    q_{y}=\sin{(\theta/2)}\cdot \cos{\beta} 
\end{equation}
\begin{equation}
    q_{z}=\sin{(\theta/2)}\cdot \cos{\gamma} 
\end{equation}
\begin{equation}
    q_{w}=\cos{(\theta/2)} 
\end{equation}
where $\theta$ represents the rotation angle transitioning from the initial position to the current position\footnotemark[1], also serves as the composition of rotations by angles $\alpha$, $\beta$ and $\gamma$ in 3D coordinate system.

\subsubsection{SKI}
In the Dynamic Skeletal Interpolation Algorithm, the Random Variation function $\mathcal{V}(\cdot)$ is used to generate a series of variants for each animation to ensure the diversity of each action, which can be denoted as following:
\begin{equation}
    \mathcal{V}(\textbf{A}, V)=\{ \mathcal{A}_v(\textbf{q}_{b}^{f}+\delta) | v\in V, b\in B, f \in F\}
\end{equation}
\begin{equation}
    \delta = U(a_{q}, b_{q})
\end{equation}
In this context, $\textbf{A}$ represents the original animation, while $V$ signifies the number of variants created.
The function $\mathcal{A}_v(\cdot)$ denotes the variant $v$ of the animation, characterized by the variant quaternion $\textbf{q}_{b}^{f}$ for each bone joint $b$ in frame $f$. 
Here, $B$ corresponds to the total number of bone joints in the animation, and $F$ indicates the total number of frames within that animation.

\subsubsection{Secene Customization}
\begin{figure}[ht]
\begin{center}
\centering
\includegraphics[width=\columnwidth]{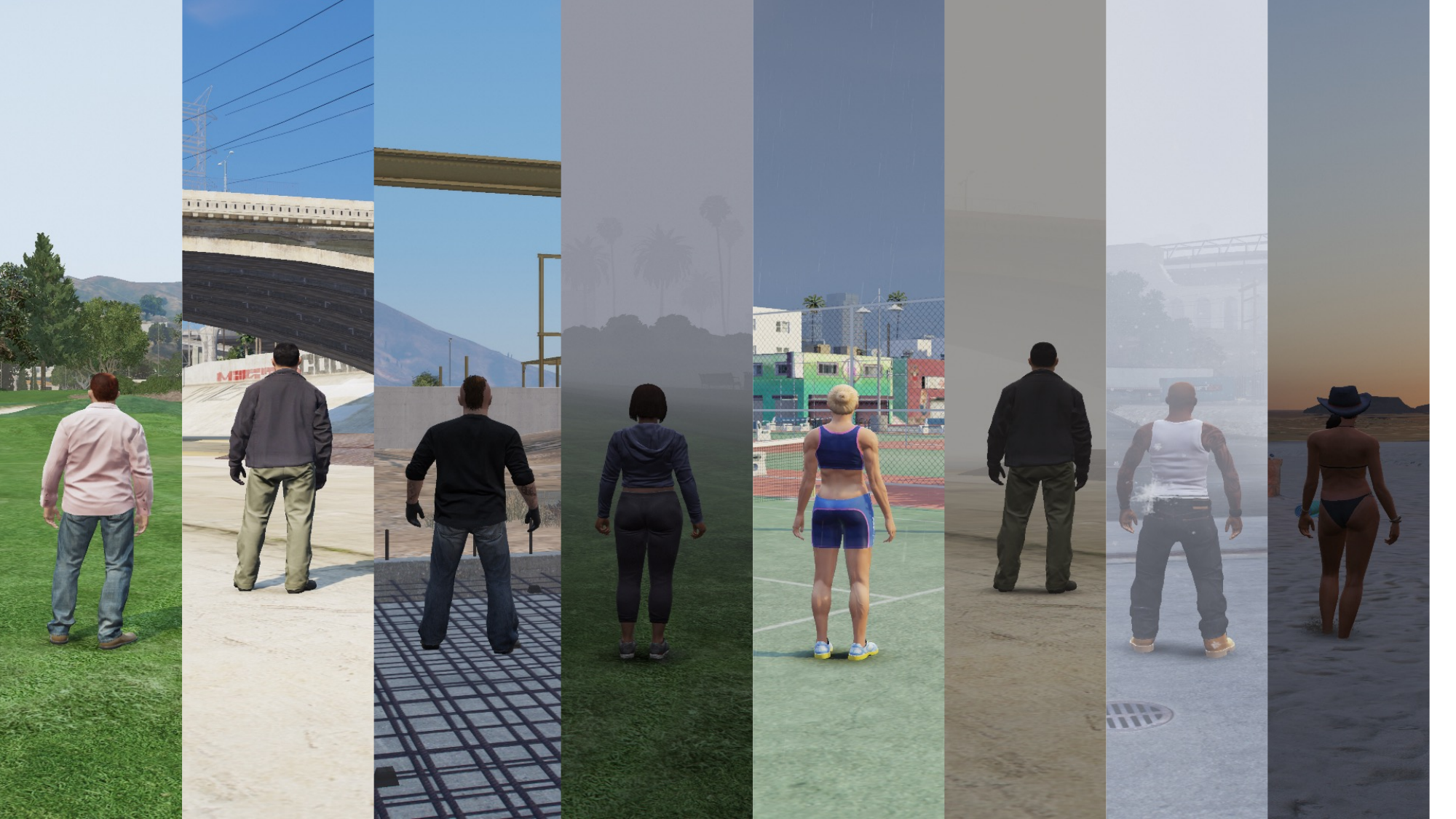} 
\end{center}
\caption{Examples of customized scenes in FiveM.}
\label{scene_customization}
\end{figure}

Environmental customization involves altering weather conditions, time of day, and in-game locations. 
Through FiveM scripts, we facilitate random weather variations, time adjustments, and repositioning across different in-game locales to achieve varied scenarios, which is shown in Figure~\ref{scene_customization}.

\begin{figure}[ht]
\begin{center}
\centering
\includegraphics[width=\linewidth]{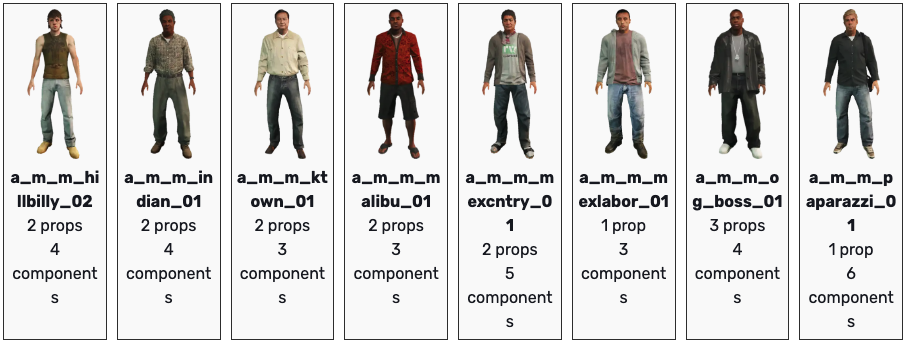}
\end{center}
\caption{Examples of native Peds.}
\label{ped}
\end{figure}

Character customization facilitates modifying both native FiveM characters and player-created avatars, enabling the simulation of actions by individuals in diverse ages, genders, and professions, reflecting the variety found in real-world scenarios.
Examples of native Peds are illustrated in Figure~\ref{ped}.

\begin{figure}[ht]
\begin{center}
\centering
\includegraphics[width=\linewidth]{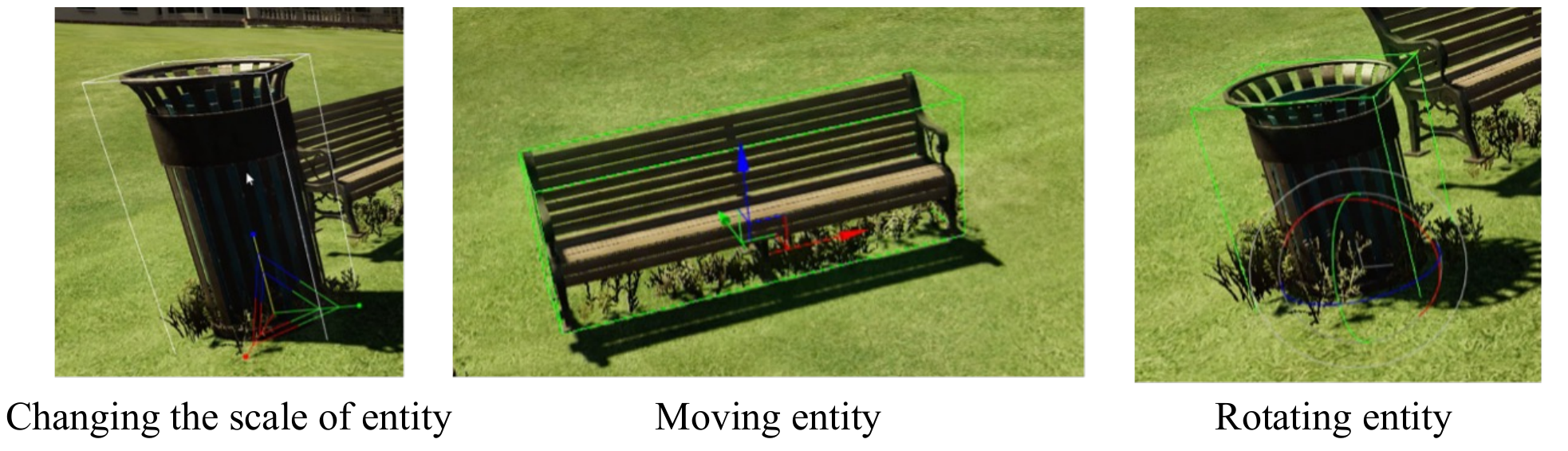}
\end{center}
\caption{Adjusting entity in CodeWorker.}
\label{codeworker}
\end{figure}

We utilize CodeWorker~\cite{codewroker}, a versatile tool designed for exploring, modifying, and exporting map data in GTAV and FiveM servers. 
This tool enables us to efficiently edit maps by adjusting various entities within the game. 
The process and capabilities of CodeWorker in facilitating these modifications are illustrated in Figure~\ref{codeworker}.
Additionally, CodeWorker offers functionality to import self-created, customized entities that have been developed using game engines.

\section{Experiment Configurations}

This section provides comprehensive details about the experimental setups mentioned in the main paper.

\subsection{Experimental Environments}
This section introduces the experimental environments used for the evaluations.
\begin{itemize}
    \item \textbf{Wisteria/BDEC-01 Supercomputer System}:\\
    CPU: Intel$^\circledR$ Xeon$^\circledR$ Gold 6258R CPU @2.70GHz;\\
    CPU number: 112;\\
    Memory: 512 GiB;\\
    GPU: NVIDIA A100-SXM4-40GB;\\
    GPU number: 8;\\
    System: Red Hat Enterprise Linux 8;\\
    Python: 3.8.18; \\
    NVCC: V11.6.124;\\
    GCC: 8.3.1 20191121 (Red Hat 8.3.1-5);\\
    PyTorch: 1.13.1;\\
    CUDA: 11.6;\\
    CuDNN: 8.3.2.\\ 

    \item \textbf{Server}:\\
    CPU: Intel$^\circledR$ Xeon$^\circledR$ W-2125 CPU @4.00GHz;\\
    CPU number: 8;\\
    Memory: 128 GB;\\
    GPU: Quadro RTX 6000;\\
    GPU number: 1;\\
    System: Ubuntu 20.04.6 LTS;\\
    Python: 3.7.16; \\
    NVCC: V11.3.58;\\
    GCC: (Ubuntu 9.4.0-1ubuntu1~20.04.2) 9.4.0;\\
    PyTorch: 1.11.0+cu113;\\
    CUDA: 11.3;\\
    CuDNN: 8.2.\\ 

    \item \textbf{PC}:\\
    CPU: Intel63 Family 6 Model 158 Stepping 9 GenuineIntel @3.6GHz;\\
    CPU number: 1;\\
    Memory: 32 GB;\\
    GPU: NVIDIA GeForce RTX 3060;\\
    GPU number: 1;\\
    System: Microsoft Windows 10.0.19045;\\
    Python: 3.8.17; \\
    NVCC: V11.8.89;\\
    GCC: N/A;\\
    PyTorch: 2.0.0;\\
    CUDA: 11.8;\\
    CuDNN: 8.7.\\ 

\end{itemize}

\subsection{Implementation Details}
This section details the implementation specifics for the various tasks outlined in the main paper.
\begin{itemize}
    \item \textbf{Wholebody 2D Pose Estimation}
    \begin{itemize}
        \item \textbf{HRNet}:\\
        Model type: Top-down;\\
        Configruation: W48;\\
        Epoch: 210;\\
        Dataset: COCO-Wholebody;\\
        Method: Heatmap;\\
        Optimizer: Adam;\\
        Learning rate: 5e-4;\\
        Warmup strategy: LinearLR;\\
        Scheduler: MultiStepLR;\\
        Input size: 384$\times$288.\\
    \end{itemize}

    \item \textbf{Action Recognition}:
    \begin{itemize}
        \item \textbf{VideoMAE}:\\
        Model type: Recognizer3D;\\
        Backbone: VisionTransformer;\\
        Frame sampling strategy: 16$\times$4$\times$1;\\
        Epoch: 150;\\
        FLOPs: 180G;\\
        Dataset type: Kinetics-400;\\
        Input size: 224$\times$224.\\

        \item \textbf{TANet}:\\
        Method: RGB;\\
        Backbone: ResNet50;\\
        Pretrain: ImageNet;\\
        Frame sampling strategy: dense-1$\times$1$\times$8;\\
        Epoch: 150;\\
        FLOPs: 43.0G;\\
        Dataset type: Kinetics-400;\\
        Input size: 224$\times$224.\\

        \item \textbf{TPN}:\\
        Method: RGB;\\
        Backbone: ResNet50;\\
        Pretrain: ImageNet;\\
        Frame sampling strategy: 8$\times$8$\times$1;\\
        Epoch: 150;\\
        Testing protocol: 10 clips$\times$3 crop;\\
        Dataset type: Kinetics-400;\\
        Input size: 320$\times$320.\\

        \item \textbf{X3D}:\\
        Method: RGB;\\
        Backbone: X3D-M;\\
        Frame sampling strategy: 13$\times$6$\times$1;\\
        Epoch: 100;\\
        Dataset type: Kinetics-400;\\
        Input size: 224$\times$224.\\

        \item \textbf{I3D}:\\
        Method: RGB;\\
        Backbone: ResNet50;\\
        Pretrain: ImageNet;\\
        Frame sampling strategy: 32$\times$2$\times$1;\\
        Epoch: 100;\\
        FLOPs: 43.5G;\\
        Dataset type: Kinetics-400;\\
        Input size: 224$\times$224.\\

        \item \textbf{I3D NL}:\\
        Method: RGB;\\
        Backbone: ResNet50 (NonLocalEmbedGauss);\\
        Pretrain: ImageNet;\\
        Frame sampling strategy: 32$\times$2$\times$1;\\
        Epoch: 100;\\
        FLOPs: 59.3G;\\
        Dataset type: Kinetics-400;\\
        Input size: 224$\times$224.\\
    \end{itemize}
\end{itemize}

\subsection{Datasets}
This section offers in-depth configurations for the NTU and H36M-based datasets that were utilized in our evaluation processes.

\subsubsection{NTU-based datasets}
Table~\ref{table:NTU-conf} presents the number of videos and frames for each dataset based on NTU.
\begin{table}
\begin{center}
\resizebox{0.5\textwidth}{!}{
\begin{tabular}{c c c}
\toprule[2pt]
\textbf{Dataset} & \textbf{Video Number} & \textbf{Frame Number} \\
\toprule[1pt]
\textbf{NTU-Original} & 1,911 & 173,129 \\
\textbf{NTU-4A} & 18,792 & 953,570 \\
\bottomrule[2pt]
\end{tabular}
}
\end{center}
\caption{Configurations of NTU-based datasets.}
\label{table:NTU-conf}
\end{table}

\subsubsection{H36M-based datasets}
Table~\ref{table:H36M-conf} presents the number of videos and frames for each dataset based on H36M.
Table~\ref{class} presents the action classes of each H36M-based dataset.
\begin{table}
\begin{center}
\resizebox{0.5\textwidth}{!}{
\begin{tabular}{c c c}
\toprule[2pt]
\textbf{Dataset} & \textbf{Video Number} & \textbf{Frame Number} \\
\toprule[1pt]
\textbf{H36M-Original} & 480 & 1,302,398 \\
\textbf{H36M-Single} & 120 & 325,419 \\
\textbf{H36M-Extracted} & 432 & 80,000 \\
\textbf{H36M-SingleExtracted} & 108 & 20,000 \\
\textbf{H36M-4A} & 3,384 & 872,564 \\
\bottomrule[2pt]
\end{tabular}
}
\end{center}
\caption{Configurations of H36M-based datasets.}
\label{table:H36M-conf}
\end{table}

\begin{table}
\begin{center}
\begin{tabular}{c  l}
\toprule[2pt]
\textbf{Label} ID & \textbf{Action}  \\
\toprule[1pt]
0 & Directions\\
1 & Discussion\\
2 & Eating\\
3 & Greeting\\
4 & Phoning\\
5 & Posing\\
6 & Purchases\\
7 & Sitting\\
8 & Sitting down\\
9 & Smoking\\
10 & Taking photo\\
11 & Waiting\\
12 & Walking \\
13 & Walking dog\\
13 & Walking together\\
\bottomrule[2pt]
\end{tabular}
\end{center}
\caption{Action classes in H36M-based datasets}
\label{class}
\end{table}

\section{Additional Results}
This section presents supplementary results from the experiments conducted in the main paper. 
It includes additional qualitative findings and detailed confusion matrices for a more comprehensive analysis.

\subsection{Additional Qualitative Results}
The extra qualitative results of the representations created by 4A are included in the supplementary materials.

\subsection{Confusion Matrices}
\begin{figure*}
\begin{center}
\includegraphics[width=\textwidth]{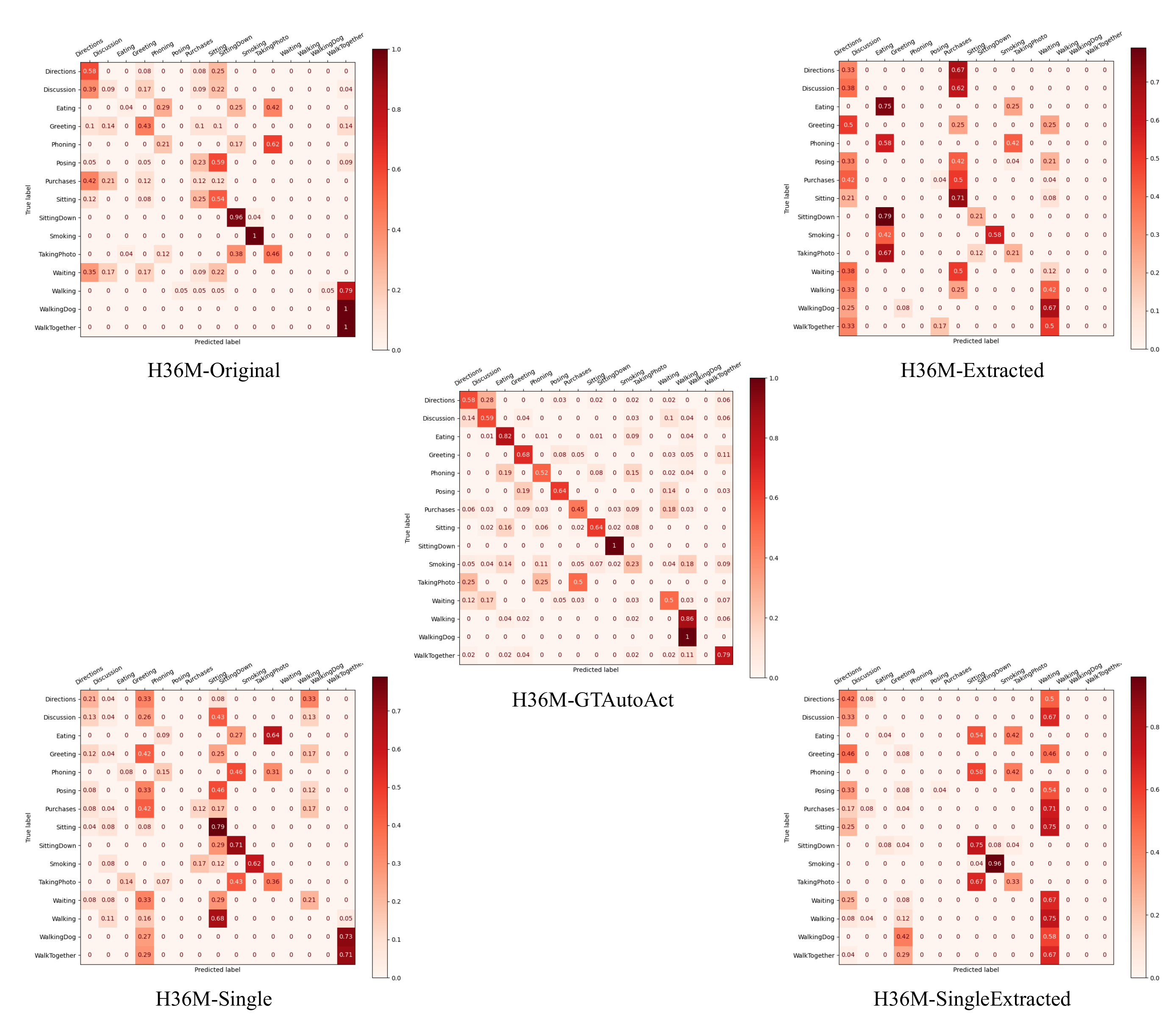}
\end{center}
   \caption{Confusion matrices of \textbf{TPN} model trained by each H36M-based datasets, tested by \textbf{H36M-Original-Test}.}
\label{cm_ori}
\end{figure*}

\begin{figure*}
\begin{center}
\includegraphics[width=\textwidth]{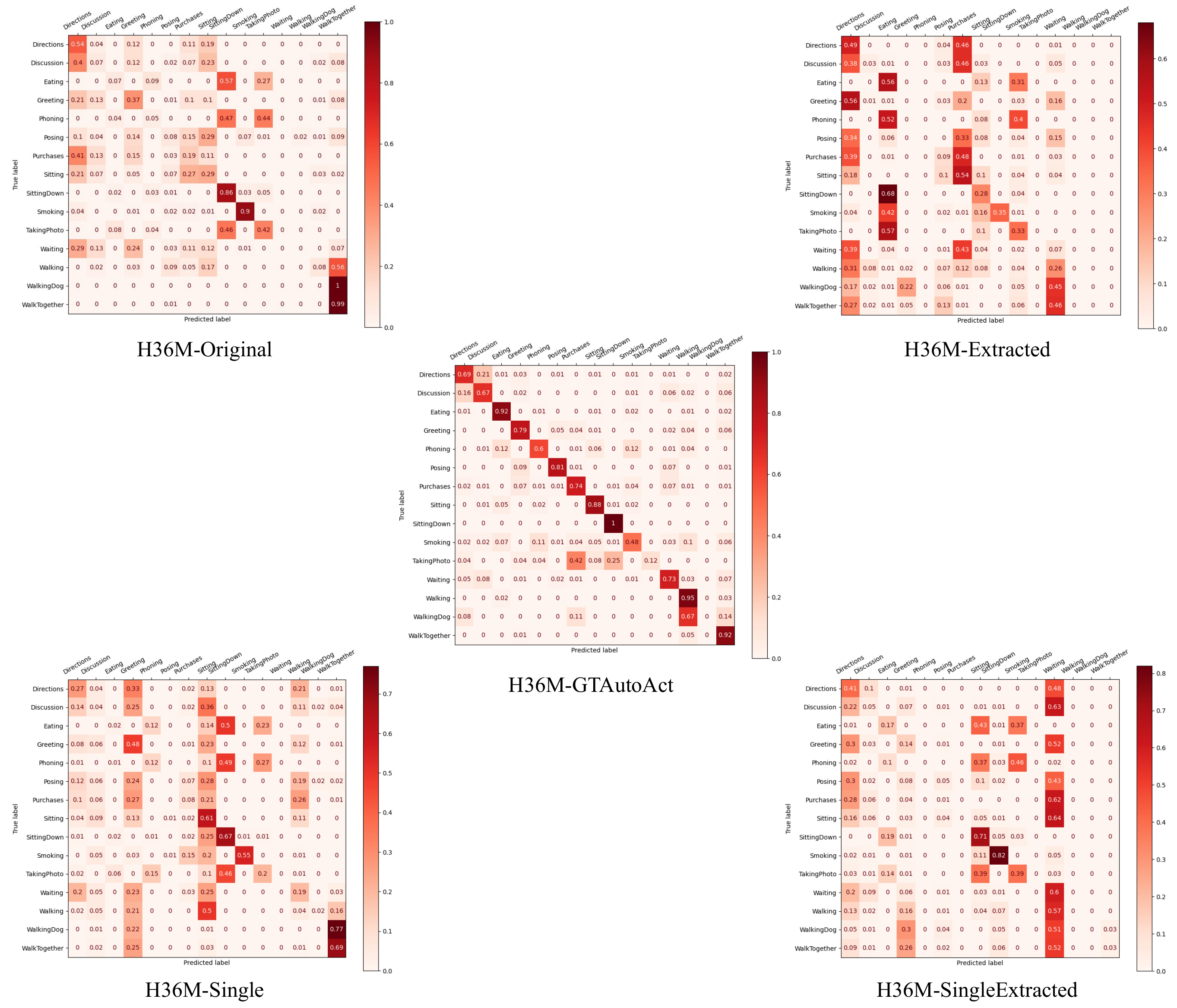}
\end{center}
   \caption{Confusion matrices of \textbf{TPN} model trained by each H36M-based datasets, tested by \textbf{H36M-Segment-Test}.}
\label{cm_seg}
\end{figure*}

This section includes selected confusion matrices for the TPN model trained on each H36M-based dataset. 
The performance of these models is evaluated using two different test datasets: H36M-Original-Test, with results displayed in Figure~\ref{cm_ori}, and H36M-Segment-Test, with results shown in Figure~\ref{cm_seg}.

\label{sec:rationale}
\end{document}